\documentclass[11pt]{article}

\usepackage{acl}
\usepackage{booktabs}
\usepackage{multirow}

\usepackage{times}
\usepackage{amssymb}
\usepackage{latexsym}
\usepackage{xcolor}

\usepackage[T1]{fontenc}
\usepackage[utf8]{inputenc}
\usepackage{microtype}
\usepackage{inconsolata}
\usepackage{graphicx}
\usepackage{fontawesome}

\usepackage[most]{tcolorbox}

\newcounter{observation}
\newtcolorbox{observationbox}{
  colback=blue!3,
  colframe=blue!65!black,
  fonttitle=\bfseries,
  coltitle=white,
  title=Observation~\theobservation,
  before upper={\stepcounter{observation}},
  arc=6pt,
  boxrule=1pt,
  left=10pt,
  right=10pt,
  top=8pt,
  bottom=8pt
}

\newtcolorbox{promptbox}[1]{
  colback=white,
  colframe=black,
  width=\columnwidth,
  boxrule=0.8pt,
  arc=4pt,
  fonttitle=\bfseries,
  breakable,
  title=\texttt{#1}
}

\title{How Long Reasoning Chains Influence LLMs' Judgment of Answer Factuality}

\author{
Minzhu Tu\textsuperscript{\rm 1,2,4} \thanks{~~Work done during an internship at ICT,CAS} \footnotemark[2] \quad
Shiyu Ni\textsuperscript{\rm 1,2,3} \thanks{~~Equal contributions} \quad
\textbf{Keping Bi}\textsuperscript{\rm 1,2,3} \thanks{~~Corresponding author} \\
\textsuperscript{\rm 1} State Key Laboratory of AI Safety \\
\textsuperscript{\rm 2} Institute of Computing Technology, Chinese Academy of Sciences \\
\textsuperscript{\rm 3} University of Chinese Academy of Sciences \\
\textsuperscript{\rm 4} Beijing University of Post and Telecommunications \\ 
\texttt{Epiphany\_1104@bupt.edu.cn} \quad
\texttt{\{nishiyu23z,bikeping\}@ict.ac.cn} \\
\href{https://github.com/Trustworthy-Information-Access/Reasoning-Affects-LLM-Judge}{\faGithub~Code} 
}

\begin{document}
\maketitle
\begin{abstract}
Large language models (LLMs) has been widely adopted as a scalable surrogate for human evaluation, yet such judges remain imperfect and susceptible to surface-level biases. One possible reason is that these judges lack sufficient information in assessing answer correctness. With the rise of reasoning-capable models, exposing a generator's reasoning content to the judge provides richer information and is a natural candidate for improving judgment accuracy. However, its actual impact on judge behavior remains understudied. In this paper, we systematically investigate how access to reasoning chains affects LLM-based judgment across factual question answering (QA) and mathematical reasoning benchmarks. We find that weak judges are easily swayed by reasoning presence, frequently accepting incorrect answers accompanied by fluent reasoning, while strong judges can partially leverage reasoning as informative evidence. Nevertheless, even strong judges are misled by seemingly high-quality reasoning chains. Controlled experiments further reveal that both fluency and factuality of reasoning chains are critical signals driving judge decisions. These findings highlight the need for more robust LLM judges that can distinguish genuine reasoning quality from superficial fluency when evaluating modern reasoning models. \looseness=-1
\end{abstract}

\section{Introduction}
Reliable evaluation is fundamental to understanding the capabilities of AI models and guiding their future development. Without an accurate assessment, identifying model strengths and limitations becomes challenging. For open-ended generation tasks, human evaluation is widely regarded as the gold standard, as it can flexibly assess semantic correctness, factuality, and overall response quality. However, human judgment is costly, time-consuming \citep{brown2020language,manas2024improving}, and difficult to scale \citep{chiang-lee-2023-large}, which limits its use in large-scale experiments and rapid model iteration. With the rapid advancement of large language models (LLMs), recent studies \citep{zheng2023judging,liu-etal-2023-g,verga2024replacing,huang-etal-2024-chatgpt,pavlovic-poesio-2024-effectiveness,tan2025judgebench} show that LLMs can deliver reference-free evaluations that closely align with human judgments, motivating their growing adoption as scalable surrogates for human evaluation in open-ended settings.

Despite their growing adoption, LLM-based judges remain imperfect. Prior studies~\citep{chen2025safer,marioriyad2025silent} have shown that LLM judgments can be sensitive to surface-level features, such as answer length, fluency, or phrasing, and may struggle to reliably distinguish correct answers from plausible but incorrect ones. One possible reason is that with only one generated answer, the judge lacks sufficient information to accurately determine its correctness. 

Recent advances have endowed LLMs with reasoning capabilities, enabling them to produce more accurate answers through explicit step-by-step thinking processes. Beyond the final answer, these visible reasoning traces offer LLM judges a richer signal for evaluation — yet whether exposing a model's reasoning process actually improves judgment quality remains an open question.
Inspired by human decision-making, we further ask whether models differ in how they use such reasoning. Humans with limited expertise may be persuaded by fluent but incorrect explanations, whereas experts can leverage reasoning as evidence to scrutinize correctness. We suspect that weak LLM judges may over-trust the presence of reasoning, while the strong ones may be better positioned to interpret reasoning as informative evidence rather than persuasive signals.

To answer these questions, we conduct a systematic study of LLM-based judgment under two settings: judge-answer-only and judge-answer-and-reasoning. For each question, we first prompt a generator model to produce a step-by-step reasoning process followed by a final answer. An LLM judge is then tasked with assessing answer quality. Under the judge-answer-only setting, the judge has access only to the final answer, whereas under judge-answer-and-reasoning, it additionally has access to the underlying reasoning process.
For the generator, we use representative open-source models from the Qwen3 series (8B, 14B, and 32B), along with the strong closed-source model DeepSeek-V3.1. As judges, we employ three series of open-source models: Qwen3, Llama 3, and GLM-4—as well as three strong closed-source models: GPT-4o, Claude Sonnet 4.5, and DeepSeek-V3.1.
To investigate the impact of task types, we conduct experiments on two factual QA datasets (NQ and HotpotQA) and two reasoning-intensive mathematical datasets (GSM8K and MATH500).

Our results show that the presence of reasoning substantially alters judgment behavior. Across all datasets, weak judges are significantly more likely to label answers as correct when reasoning is provided, even when those answers are incorrect. In contrast, strong judges exhibit more selective behavior. They are not merely swayed by the reasoning, but can, in some cases, identify errors within it and use the reasoning content to more effectively assess correctness.
However, all models show an increased tendency to judge incorrect answers as correct after being exposed to reasoning chains generated by DeepSeek-V3.1. This suggests that even strong judges can be misled by seemingly high-quality chains of reasoning.

Given the complexity of natural reasoning chains, we conduct controlled experiments to isolate the effects of two key attributes—fluency and factuality—by manipulating them and observing their impact on the judge’s decisions. Specifically, we disrupt fluency by inserting factually correct but irrelevant knowledge into otherwise coherent reasoning chains, and degrade factuality by replacing such knowledge with counterfactual information. 
Disrupting the fluency makes nearly all judges more likely to label the answer as incorrect, an effect that is further amplified when counterfactual content is introduced. These findings suggest that both fluency and factuality of the reasoning chain serve as critical signals in LLM-based judgment.
Overall, these findings highlight the need to move beyond answer-only evaluation paradigms and call for more robust judge designs that can critically assess, rather than passively consume, the reasoning processes of modern LLMs.

\begin{figure*}[t]
  \centering
  \includegraphics[width=\textwidth]{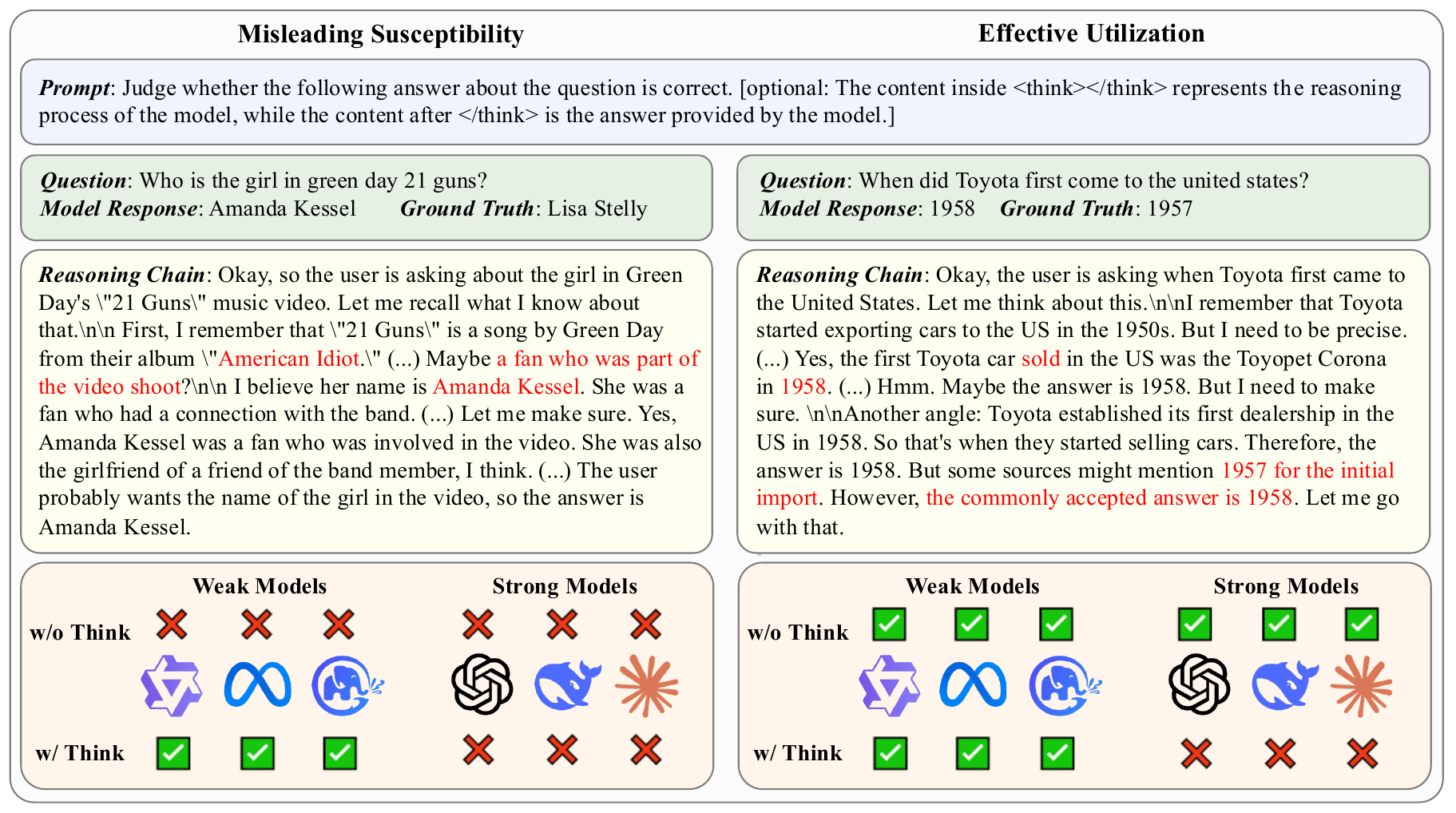}
  \caption{Examples on how reasoning chains affect LLM-based judgment. These are two question-answering examples from NQ, where answers and reasoning processes are generated by Qwen3-8B. “w/o Think” means the judge evaluates correctness based only on the model’s answer, while “w/ Think” means the judge can see the model’s reasoning process when assessing answer accuracy. The left part shows that weak judges are misled by reasoning that appears fluent but is actually incorrect after reviewing the reasoning process. The right figure shows that strong judges identify errors in the reasoning and successfully assess the correctness of the answer.}

  \label{fig:framework}
\end{figure*}

\section{Related Work}
\subsection{LLM-as-a-Judge}
The rapid advancement of LLMs has expanded their utility beyond traditional text generation tasks, owing to their strong performance across diverse tasks~\citep{shi2025deep}. Because traditional automatic metrics often fail to capture the quality and reasoning depth of open-ended outputs, the LLM-as-a-Judge paradigm has emerged as a scalable alternative to human evaluation. Early work such as MT-Bench \citep{zheng2023judging} and Chatbot Arena \citep{chiang2024chatbot} demonstrated that strong models like GPT-4 can produce judgments that correlate well with human preferences. Further research, such as G-Eval \citep{liu-etal-2023-g} and AlpacaFarm \citep{dubois2023alpacafarm}, is viewed as a multi-aspect scoring or pairwise comparison task, leveraging chain-of-thought (CoT) reasoning to enhance the interpretability and reliability of automated judgments.

Despite its effectiveness, prior work has identified several limitations of LLM-based judges, including positional bias, sensitivity to input perturbations, and self-conflicting evaluations \citep{wang2024large}. To address these issues, recent efforts focus on developing specialized evaluators and more robust evaluation protocols. For example, PandaLM \citep{wang2023pandalm}, Prometheus \citep{kim2023prometheus}, and JudgeLM \citep{zhu2023judgelm} are fine-tuned to provide fine-grained assessments, while Auto-J \citep{li2023generative} enables flexible evaluation generation. In addition, approaches such as multi-agent debate \citep{du2024improving,chan2023chateval} and iterative self-refinement \citep{madaan2023self} have been explored to further enhance the factuality and consistency of judgments, ensuring a more principled and trustworthy evaluation discipline for the next generation of language models.


\subsection{Reasoning Models}
Recent advancements enable LLMs to generate intermediate steps, commonly termed reasoning chains, prior to producing a final answer. A prominent approach in this domain is Chain-of-Thought (CoT) prompting \citep{wei2022chain,kojima2022large}, which elicits reasoning chains in natural language and has shown to be effective in enhancing performance across complex reasoning tasks. To enhance the reliability of these reasoning chains, self-consistency \citep{wang2022self} introduces a majority-voting mechanism, while least-to-most prompting \citep{zhou2022least} aims to narrow the compositionality gap by decomposing complex problems into independent sub-problems \citep{press2023measuring}. Furthermore, research on self-taught reasoner \citep{zelikman2022star} demonstrates that models can bootstrap their reasoning abilities by iteratively generating rationales and fine-tuning on correct solutions.
Beyond these linear paradigms, recent work explores more structured and iterative reasoning. Frameworks such as tree of thoughts \citep{yao2023tree} and graph of thoughts \citep{besta2024graph} allow models to explore multiple reasoning branches and backtrack using search algorithms. Additionally, agent-based methods like reflexion \citep{shinn2023reflexion} introduce iterative feedback and refinement. These advances not only improve reasoning performance, but also make reasoning chains more structured and informative. Prior study \citep{zhang2025reasoning} suggests that such chains encode information indicative of model reliability. 

As additional information beyond the answer itself, chain-of-thought reasoning, as a potential means of improving the judgment accuracy of LLM-as-a-judge, has not yet been studied in terms of its impact on LLM-as-a-judge. In this work, we investigate how reasoning chains influence judging behaviors when LLMs are used as evaluators and provide a systematic experimental analysis.

\section{Task Formulation}

\paragraph{Reasoning-enhanced Question Answering.}
Let $q$ denote a question.  
A generator model $G$ produces a reasoning chain before arriving at the final answer:
\begin{equation}
(r, a) = G(q),
\end{equation}
where $r$ represents the reasoning process (e.g., content between \texttt{<think>} and \texttt{</think>} ), and $a$ is the final answer.
We denote the ground-truth answer as $a^*$. The correctness of the generated answer is defined as:
\begin{equation}
y = \mathbb{I}(a = a^*),
\end{equation}
where $y \in \{0,1\}$.

\paragraph{LLM-as-a-Judge.}
Given a question $q$ and a generated answer $a$, 
a judge model $J$ is asked to evaluate whether the answer is correct. 
The judge outputs a binary decision:
\begin{equation}
\hat{y} = J(q, a) \in \{0,1\},
\end{equation}
where $\hat{y} = 1$ indicates that $J$ consider the answer $a$ is correct, and $\hat{y} = 0$ otherwise.
We refer to this as \textit{judging without reasoning}.

Since the reasoning process provides more signals for evaluation, in this paper, we introduce \textit{judging with reasoning}. The judge model determines whether the result is correct based on the question, the answer, and the reasoning process.
\begin{equation}
\hat{y}^{\text{reason}} = J(q, a,r) \in \{0,1\},
\end{equation}
By comparing $\hat{y}$ and $\hat{y}^{\text{reason}}$, we can isolate the effect of reasoning on the LLM judge:
\begin{equation}
\Delta J = \hat{y}^{\text{reason}} - \hat{y}.
\end{equation}

\section{Experimental Setup}
\subsection{Models}
For the answer generation models, we use representative reasoning-enhanced LLMs, including the Qwen3~\citep{yang2025qwen3} series models (8B, 14B, 32B) as well as DeepSeek-V3.1~\citep{deepseekai2025deepseekv3technicalreport}.
For the judge models, we employ models with a range of capabilities to examine whether the intrinsic ability of the judge model is correlated with the influence of reasoning processes. Specifically, we used open-source models including the Qwen3 series (8B, 14B, 32B), Llama-3.1 (8B and 70B)~\citep{meta_llama31_modelcard}, GLM-4-32B~\citep{zai_glm4_32b}, GLM-4-Z1-32B~\citep{zai_glm4_32b}, and DeepSeek-V3.1, as well as two closed-source models from the most powerful tier: GPT-4o \citep{openai2024gpt4ocard} and Claude Sonnet 4.5 \citep{anthropic_claude_opus_4_5_2025}. 
We consider DeepSeek-V3.1 and the two closed-source models as strong models based on their model size and QA capabilities (See Figure~\ref{fig:Acc_Compare}), while the remaining models are considered weak models.

\subsection{Datasets}
We comprehensively evaluate the impact of reasoning content on LLM-as-a-judge across two factual datasets and two reasoning-intensive mathematical datasets. The factual QA datasets include Natural Questions (NQ)~\citep{kwiatkowski2019natural}, which consists of single-hop factual questions, and HotpotQA~\citep{yang2018hotpotqa}, which focuses on question that require multi-hop reasoning. The mathematical datasets include GSM8K~\citep{cobbe2021training}, a collection of grade-school-level math word problems requiring multi-step reasoning, and MATH-500~\citep{hendrycks2021measuring}, which comprises more challenging, competition-level problems designed to assess advanced mathematical reasoning.
To manage the computational costs associated with closed-source models, we randomly sample 500 questions from each dataset.

\subsection{Evaluation Metrics}
We use accuracy to measure QA performance, defined as the proportion of generated responses that match the ground-truth labels. To ensure a reliable assessment of correctness, we employ Qwen2.5-72B-Instruct~\citep{qwen2025qwen25technicalreport} to verify the consistency between model-generated answers and ground-truth answers.
Following prior work~\citep{ni2024llms,ni2025towards}, we further evaluate the performance of judge models using four metrics: 1) Alignment, the proportion of cases where the judge’s verdict agrees with the ground-truth correctness; 2) Pass Rate, the proportion of cases where the judge deems an answer correct; 3) Overconfidence, the proportion of incorrect answers that are mistakenly judged as correct; and 4) Conservativeness, the proportion of correct answers that are incorrectly judged as incorrect.

\subsection{Implementation Details}
We prompt the generator to produce an output for each question, consisting of both a reasoning chain and a final answer. Judges are instructed to assign a binary score. We use a temperature of 0.6 during inference. Additional details are provided in Appendix~\S~\ref{sec:prompts}.

\section{Results and Analysis}

\begin{table*}[t]
\centering
\caption{Evaluation results (\%) of LLM-as-a-Judge behavior with and without reasoning chains across factual and mathematical datasets, with all answers generated by Qwen3-8B. Bolds denote the highest on each dataset.}
\scriptsize

\setlength{\tabcolsep}{6pt}
\begin{tabular}{l c l cc cc cc cc}
\toprule
\multirow{2}{*}{Dataset}
& \multirow{2}{*}{Acc}
& \multirow{2}{*}{Judge Models}
& \multicolumn{2}{c}{Alignment}
& \multicolumn{2}{c}{Pass Rate}
& \multicolumn{2}{c}{Overconfidence}
& \multicolumn{2}{c}{Conservativeness} \\
\cmidrule(lr){4-5} \cmidrule(lr){6-7} \cmidrule(lr){8-9} \cmidrule(lr){10-11}
 &  & 
& w/o Think & w/ Think
& w/o Think & w/ Think
& w/o Think & w/ Think
& w/o Think & w/ Think \\
\midrule

\multirow{10}{*}{NQ}
& \multirow{10}{*}{23.2} & Qwen3-8B      
& 58.2 & 33.2 & 57.8 & 88.0 & 38.2 & 65.8 & 3.6 & 1.0 \\
&       & Qwen3-14B     
& 58.0 & 36.4 & 58.0 & 84.0 & 38.4 & 62.2 & 3.6 & 1.4 \\
&       & Qwen3-32B     
& 41.4 & 36.8 & 79.4 & 83.6 & 57.4 & 61.8 & 1.2 & 1.4 \\
&       & Llama3-8B     
& 38.6 & 28.4 & \textbf{79.8} & \textbf{93.2} & \textbf{59.0} & \textbf{70.8} & 2.4 & 0.8 \\
&       & Llama3-70B    
& 54.0 & 41.4 & 66.4 & 79.0 & 44.6 & 57.2 & 1.4 & 1.4 \\
&       & GLM4-32B      
& 52.0 & 35.8 & 69.6 & 87.0 & 47.2 & 64.0 & 0.8 & 0.2 \\
&       & GLM4-Z1-32B   
& \textbf{79.4} & 52.8 & 32.2 & 63.6 & 14.8 & 43.8 & \textbf{5.8} & 3.4 \\
\cmidrule(lr){3-11}
&       & GPT-4o        
& 74.0 & 74.0 & 44.0 & 41.2 & 23.4 & 22.0 & 2.6 & 4.0 \\
&       & DeepSeek-v3.1 
& 63.4 & 76.2 & 55.8 & 35.4 & 34.6 & 18.0 & 2.0 & 5.8 \\
&       & Claude Sonnet 4.5   
& 75.8 & \textbf{84.0} & 43.0 & 18.8 & 22.0 & 5.8 & 2.2 & \textbf{10.2} \\

\midrule
\multirow{10}{*}{HotpotQA}
& \multirow{10}{*}{26.6} & Qwen3-8B      
& 59.2 & 44.6 & 53.4 & 78.0 & 33.8 & 53.4 & 7.0 & 2.0 \\
&       & Qwen3-14B     
& 64.0 & 47.2 & 51.4 & 76.6 & 30.4 & 51.4 & 5.6 & 1.4 \\
&       & Qwen3-32B     
& 51.6 & 47.6 & 72.6 & 76.6 & 47.2 & 51.2 & 1.2 & 1.2 \\
&       & Llama3-8B     
& 47.2 & 40.0 & \textbf{73.0} & \textbf{85.0} & \textbf{49.6} & \textbf{59.2} & 3.2 & 0.8 \\
&       & Llama3-70B    
& 58.2 & 48.4 & 62.0 & 73.8 & 38.6 & 49.4 & 3.2 & 2.2 \\
&       & GLM4-32B      
& 58.4 & 42.2 & 65.0 & 83.6 & 40.0 & 57.4 & 1.6 & 0.4 \\
&       & GLM4-Z1-32B   
& 78.0 & 59.0 & 13.8 & 48.8 & 4.6 & 31.6 & \textbf{17.4} & 9.4 \\
\cmidrule(lr){3-11}
&       & GPT-4o        
& 78.8 & \textbf{78.4} & 36.6 & 28.2 & 15.6 & 11.6 & 5.6 & 10.0 \\
&       & DeepSeek-v3.1 
& 78.6 & 76.8 & 28.0 & 13.4 & 11.4 & 5.0 & 10.0 & 18.2 \\
&       & Claude Sonnet 4.5   
& \textbf{84.6} & \textbf{78.4} & \textbf{73.0} & \textbf{85.0} & \textbf{49.6} & \textbf{59.2} & 6.0 & \textbf{21.8} \\

\midrule
\multirow{10}{*}{GSM8K}
& \multirow{10}{*}{94.0} & Qwen3-8B      
& 75.2 & 94.6 & 74.0 & 99.0 & 2.4 & 5.2 & 22.4 & 0.2 \\
&       & Qwen3-14B     
& 72.6 & 94.0 & 71.4 & \textbf{99.2} & 2.4 & \textbf{5.6} & 25.0 & 0.4 \\
&       & Qwen3-32B     
& 89.2 & 94.0 & 91.6 & \textbf{99.2} & 4.2 & \textbf{5.6} & 6.6 & 0.4 \\
&       & Llama3-8B     
& 89.0 & 94.6 & 93.0 & 99.0 & 5.0 & 5.2 & 6.0 & 0.2 \\
&       & Llama3-70B    
& 83.4 & 93.8 & 86.2 & 98.6 & 4.4 & 5.4 & 12.2 & 0.8 \\
&       & GLM4-32B      
& 87.6 & \textbf{95.0} & 91.6 & 99.0 & 5.0 & 5.0 & 7.4 & 0.0 \\
&       & GLM4-Z1-32B   
& 44.0 & 83.8 & 42.8 & 86.2 & 2.4 & 4.2 & \textbf{53.6} & 12.0 \\
\cmidrule(lr){3-11}
&       & GPT-4o        
& 91.0 & 92.4 & 92.6 & 94.4 & 3.8 & 4.0 & 5.2 & 3.6 \\
&       & DeepSeek-v3.1 
& \textbf{93.2} & 94.4 & \textbf{96.8} & \textbf{99.2} & \textbf{5.0} & \textbf{5.6} & 2.0 & 0.4 \\
&       & Claude Sonnet 4.5   
& 66.0 & 70.0 & 63.6 & 68.4 & 1.8 & 2.2 & 32.2 & \textbf{27.8} \\

\bottomrule
\end{tabular}
\label{tab:Qwen3-8B_judge}
\end{table*}

\subsection{General Results}
Table~\ref{tab:Qwen3-8B_judge} presents the evaluation results using Qwen3-8B as the generator, assessing a range of judge models across multiple datasets. Results with other generators (e.g., DeepSeek-v3.1) are deferred to Appendix~\ref{sec:Other_Generator} due to space constraints. We observe that the impact of exposing reasoning chains to judge models varies systematically with the capability of the judge model.
\begin{observationbox}
Weak judges tend to be misled by reasoning chains, resulting in inflated pass rates.
\end{observationbox}
For weak judges such as Qwen3-8B, the exposure of reasoning chains significantly increases pass rate and harms alignment in most cases. For example, on NQ, although only 23.2\% of the answers are correct, Qwen3-8B exhibits a high pass rate of 57.8\%, indicating substantial overconfidence. Moreover, when the model is exposed to chain-of-thought reasoning, the pass rate increases further. This suggests that weaker judges can be misled by the content of the reasoning chain.

A similar trend is also observed on the mathematical dataset, where weak judges continue to exhibit an increased pass rate after the exposure of reasoning chains. On GSM8K, nearly all weak judges show a substantial increase in pass rate after being exposed to chain-of-thought reasoning, in some cases approaching 100\%. Since GSM8K is relatively simple—with about 94\% of the answers being correct—this can create the illusion of improved alignment. However, on the more challenging MATH dataset (See Table~\ref{tab:math500_results}), especially on our balanced subset of correct and incorrect samples (See Table~\ref{tab:math500_balanced_results}), weak judges also exhibit a significant increase in pass rate after seeing the reasoning chains. This leads to pass rates far exceeding the true correctness rate, indicating severe overconfidence.
We think that natural reasoning chains can mislead weak judges because they may remain internally coherent even when built upon an early error. For example, in Figure~\ref{fig:framework} for the question ``Who is the girl in Green Day's 21 Guns," the generator's reasoning departs at the outset by incorrectly associating 21 Guns with "American Idiot" rather than "21st Century Breakdown", which is the correct answer. Despite this incorrect premise, the subsequent reasoning forms a locally consistent and plausible narrative. 

\begin{observationbox}
Strong models exhibit more selective behavior and, in some cases, effectively leverage the provided reasoning chains to improve their judgments.
\end{observationbox}
In contrast to weak models, strong models often exhibit a decrease in pass rates after seeing the reasoning chain, and this reduction is sometimes accompanied by an improvement in alignment. For example, On NQ, DeepSeek-v3.1's alignment increases from 63.4\% to 76.2\%, while its pass rate decreases from 55.8\% to 35.4\%, indicating that the model becomes more selective rather than indiscriminately judge the answer as correct. Meanwhile, its overconfidence declines from 34.6\% to 18.0\%, suggesting that exposure to the reasoning chains enables the model to rectify its initial misconceptions, successfully identifying errors it had previously overlooked.

However, strong models do not consistently leverage reasoning information effectively. On HotpotQA, GPT-4o's pass rate decreases from 36.6\% to 28.2\%, and overconfidence drops from 15.6\% to 11.6\%, suggesting successful identification of some errors. However, its alignment remains virtually unchanged, while the conservativeness rises from 5.6\% to 10.0\%, which implies that the model begins to incorrectly reject originally correct answers. This tendency towards excessive skepticism is also observed in the NQ dataset.
Interestingly, even strong models can be misled by reasoning chains. As shown in Table~\ref{tab:DeepSeek-v3.1_judge}, all models exhibit a significant increase in pass rate after being exposed to reasoning chains generated by DeepSeek-V3.1. We hypothesize that even for incorrect answers, the reasoning chains produced by DeepSeek-V3.1 appear highly plausible and well-structured, thereby misleading all models.
These findings suggest that providing the model’s reasoning process during evaluation has the potential to improve judgment performance, but its effectiveness depends on the capability of the judge model. Current models are still unable to reliably and consistently make effective use of the reasoning process. 

\begin{observationbox}
Self-judging exhibits trends similar to those observed in the generate-then-judge setting.
\end{observationbox}
To investigate whether models exhibit a preference toward their own reasoning chains, we ask the model to judge the correctness of its answer immediately after generating it. In this setting, the model is aware that it is evaluating its own output. As shown in Table~\ref{tab:selfjudge}, self-judging exhibits pass rate patterns similar to those observed when the same model serves as an external judge, and significantly different from settings where reasoning chains are invisible. These results indicate that misleading effects arise from the presence of reasoning chains alone, independent of whether judgment is performed by a separate model.


\begin{table*}[t]
\centering
\caption{Results of self-judging experiments on the NQ dataset. Underlined values indicate the self-judging setting and the corresponding generate-and-judge setting with the same model.}
\small
\setlength{\tabcolsep}{3.5pt}
\begin{tabular}{l c l cc cc cc cc}
\toprule

\multirow{2}{*}{Generator}
& \multirow{2}{*}{Acc}
& \multirow{2}{*}{Judge Models}
& \multicolumn{2}{c}{Alignment}
& \multicolumn{2}{c}{Pass Rate}
& \multicolumn{2}{c}{Overconfidence}
& \multicolumn{2}{c}{Conservativeness} \\
\cmidrule(lr){4-5} \cmidrule(lr){6-7} \cmidrule(lr){8-9} \cmidrule(lr){10-11}
 &  & 
& w/o Think & w/ Think
& w/o Think & w/ Think
& w/o Think & w/ Think
& w/o Think & w/ Think \\
\midrule

\multirow{4}{*}{Qwen3-8B}
& \multirow{4}{*}{36.6} & Self-judge & --    & \underline{63.2} & --    & \underline{61.4} & --    & 30.8 & --    & 6.0 \\
&                        & Qwen3-8B  & 63.4 & \underline{63.3} & 53.7 & \underline{61.7} & 26.9 & 30.9 & 9.7  & 5.8 \\
&                        & Qwen3-14B & 67.3 & 63.7 & 53.3 & 62.6 & 24.7 & 31.2 & 8.0  & 5.1 \\
&                        & Qwen3-32B & 54.9 & 62.3 & 77.3 & 65.1 & 42.9 & 33.1 & 2.2  & 4.6 \\

\midrule
\multirow{4}{*}{Qwen3-14B}
& \multirow{4}{*}{41.4} & Self-judge & --    & \underline{59.8} & --    & \underline{71.5} & --    & 35.1 & --    & 5.0 \\
&                        & Qwen3-8B  & 64.3 & 61.2 & 55.0 & 69.6 & 24.6 & 33.5 & 11.1 & 5.4 \\
&                        & Qwen3-14B & 64.6 & \underline{61.3} & 60.2 & \underline{70.9} & 27.1 & 34.1 & 8.3  & 4.6 \\
&                        & Qwen3-32B & 54.4 & 59.1 & 81.9 & 74.3 & 43.0 & 36.8 & 2.6  & 4.0 \\

\midrule
\multirow{4}{*}{Qwen3-32B}
& \multirow{4}{*}{45.9} & Self-judge & --    & \underline{60.1} & --    & \underline{79.5} & --    & 36.8 & --    & 3.1 \\
&                        & Qwen3-8B  & 62.3 & 58.2 & 57.3 & 82.8 & 24.6 & 39.3 & 13.2 & 2.4 \\
&                        & Qwen3-14B & 65.5 & 58.4 & 59.9 & 83.2 & 24.3 & 39.5 & 10.2 & 2.1 \\
&                        & Qwen3-32B & 55.4 & \underline{57.9} & 85.1 & \underline{83.9} & 41.9 & 40.1 & 2.7  & 2.1 \\

\bottomrule
\end{tabular}
\label{tab:selfjudge}
\end{table*}

\subsection{Detailed Analysis of Pass Rate \label{sec:detailed_analysis}}
The results in the previous section show that exposing reasoning chains substantially increases the pass rates of weak models. To further understand what drives the change in pass rate, we examine which subsets of the data contribute to this shift. For example, for weak models, we analyze whether the overall increase arises from a trade-off—where pass rates decrease on some samples but increase on a larger number of others—or whether there are few, if any, cases in which the pass rate decreases.

\begin{figure}[t]
    \centering
    \includegraphics[width=\linewidth]{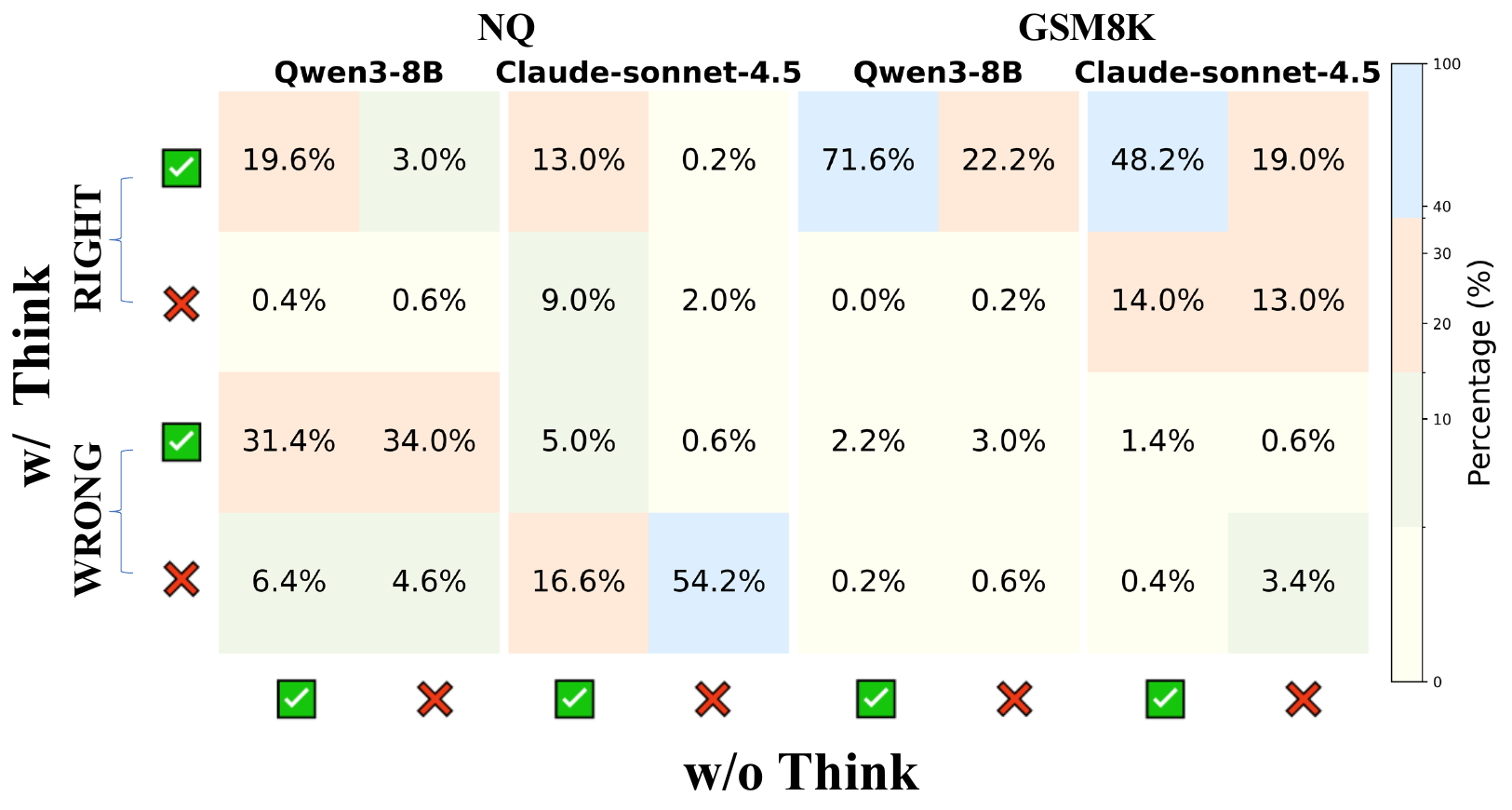}
    \caption{Distribution of judge decisions under different answer correctness and reasoning visibility. Each block shows the percentage of samples falling into a specific judgment transition across datasets and judge models. \textit{w/ Think} and \textit{w/o Think} indicate whether the reasoning chain is provided. $\checkmark$ indicates a \texttt{certain} verdict and $\times$ indicates \texttt{uncertain}, while RIGHT and WRONG refer to correct and incorrect generator answers.}
    \label{fig:pass_rate_source}
\end{figure}

As shown in Figure~\ref{fig:pass_rate_source}, more than half of the samples retain the same judgment regardless of whether the judges are strong or weak. We therefore focus on the subset of samples where the models revise their judgments after being exposed to the reasoning chains.
On NQ, for the weak judge (i.e., Qwen3-8B), the increase in pass rate is mainly driven by cases where the model initially labels an actually incorrect answer as incorrect but revises its judgment to correct after seeing the reasoning chain. This misleading effect accounts for 34.0\% of all samples and constitutes the majority of cases with changed judgments. This indicates that the increase in pass rate is largely due to models being misled by the reasoning chains.
In contrast, for Claude Sonnet 4.5, misleading reasoning accounts for only 0.6\% of cases. After reviewing the reasoning chains, the model correctly revises its judgments for 16.6\% of incorrect answers, which also makes up the majority of cases with changed judgments. This difference contributes to the higher alignment observed in strong models and reflects their lower susceptibility to misleading reasoning chains. \looseness=-1

On GSM8K, since the dataset is relatively simple and most answers are correct, the weak judge tends to label nearly all samples as correct. As a result, it appears to correctly revise its judgments on 22.2\% of samples with correct answers after being exposed to the reasoning chains. However, the proportion of misleading cases for weak models is 3.0\%, still much higher than the 0.6\% observed for the strong judge. For the strong judge, after reviewing the reasoning chains, the model incorrectly changes its judgments to incorrect on 14\% of the samples, suggesting a tendency to be overly critical and to over-reject correct answers.
Our detailed analysis further supports that the higher pass rate of weak models primarily arises from cases where incorrect answers are incorrectly accepted as correct once reasoning chains are provided. In contrast, such misclassification is substantially less frequent in strong models, indicating that weak judges are significantly more susceptible to being misled by the reasoning chains.

\begin{table*}[t]
\centering
\caption{Judge pass rates (\%) on the NQ dataset under synthesized prefix and suffix injections.}
\small
\setlength{\tabcolsep}{6pt}
\begin{tabular}{c c l cc cc cc cc}
\toprule
\multirow{2}{*}{Generator}
& \multirow{2}{*}{Acc}
& \multirow{2}{*}{Judge Models}
& \multicolumn{2}{c}{Vanilla}
& \multicolumn{2}{c}{Basic-All}
& \multicolumn{2}{c}{Wrong-Few}
& \multicolumn{2}{c}{Wrong-All} \\
\cmidrule(lr){4-5} \cmidrule(lr){6-7} \cmidrule(lr){8-9} \cmidrule(lr){10-11}
& & 
& w/o Think & w/ Think
& Prefix & Suffix
& Prefix & Suffix
& Prefix & Suffix \\
\midrule

\multirow{6}{*}{Qwen3-8B}
& \multirow{6}{*}{44.8}
& Qwen3-8B
& 63.0 & 91.0
& 61.0 & 86.0
& 58.8 & 82.2
& 47.6 & 76.2 \\

& & Qwen3-14B
& 62.8 & 87.4
& 58.0 & 81.6
& 55.0 & 76.6
& 53.2 & 67.8 \\

& & Qwen3-32B
& 79.4 & 85.0
& 74.8 & 88.8
& 73.2 & 87.8
& 71.6 & 81.8 \\

& & Llama3-8B
& 74.8 & 97.2
& 92.8 & 95.4
& 95.0 & 96.4
& 95.2 & 95.6 \\

& & GPT-4o
& 50.6 & 48.6
& 18.4 & 39.2
& 13.4 & 35.8
& 23.4 & 36.2 \\

& & DeepSeek-v3.1
& 45.2 & 24.4
& 1.0  & 1.8
& 1.0  & 1.8
& 0.0  & 10.0 \\

\bottomrule
\end{tabular}

\label{tab:synthesized_reasoning}
\end{table*}
\subsection{How Do Synthesized Reasoning Chains Affect LLM-based Judging ?}
In the previous experiments, we mainly examined how natural reasoning chains affect judge behavior. While this setting reflects practical usage, the complexity of natural reasoning chains makes it difficult to isolate the specific factors that influence model judgments. Intuitively, reasoning chains often contain content that appears fluent but is factually incorrect. Such fluent yet incorrect reasoning may mislead the model, while the model may also identify factual errors in the reasoning chain. Therefore, we design controlled experiments to investigate the effects of fluency and factuality in reasoning chains on model judgments.

To manipulate fluency, we insert fixed-length, question-irrelevant common-sense statements (e.g., The Earth orbits the Sun once every 365 days) into the reasoning chain. These statements are fluent and factually correct but irrelevant to the question, thereby disrupting the coherence of the reasoning process.
To examine factuality, we modify these statements into counterfactual variants (e.g., changing “365 days” to “100 days”). We further hypothesize that the position of the injected content may affect judgment outcomes. Therefore, we insert such question-irrelevant statements at either the beginning or the end of the reasoning chain to study the effect of their placement. 
To disrupt fluency, we insert four common-sense statements; to disrupt factuality, we progressively modify these statements into counterfactual ones. Results are shown in Figure~\ref{fig:factual_injection} and more detailed results can be found in Table~\ref{tab:fine_grained_errors}.

\subsubsection{Effects of Fluency}
\begin{figure}[t]
  \centering
  \includegraphics[width=\linewidth]{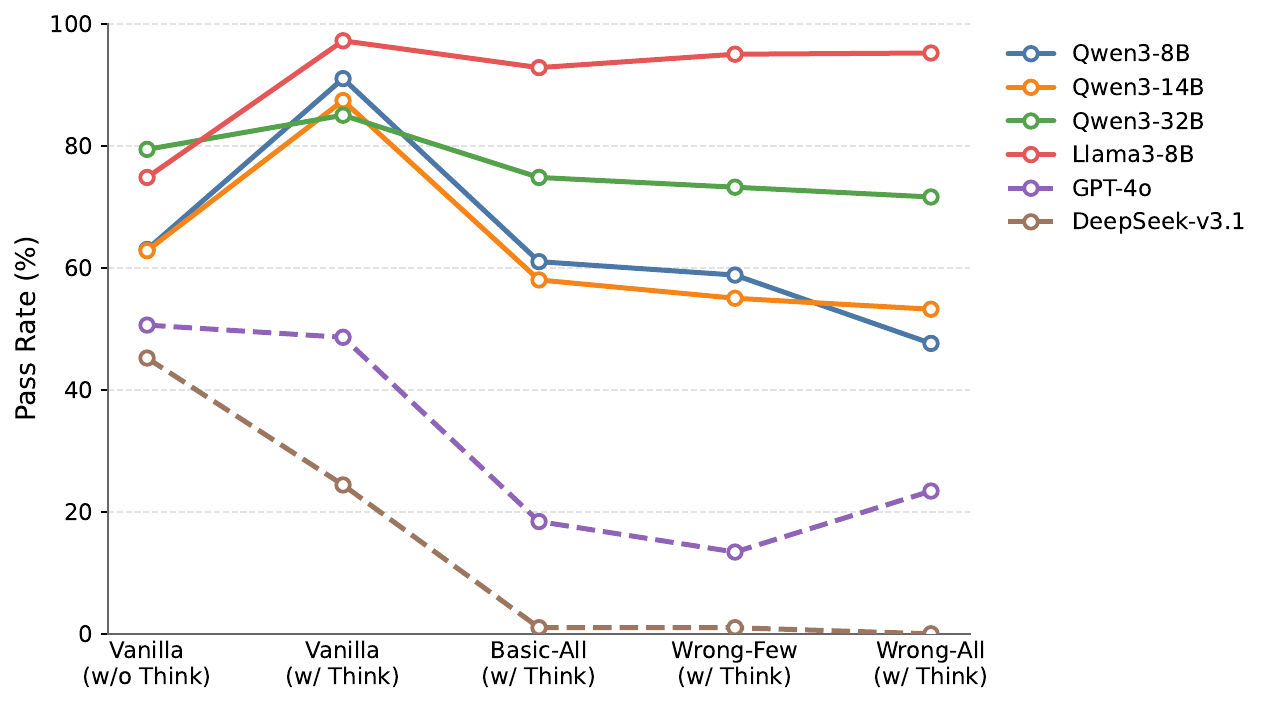}
  \caption{Pass rates under factual injections into reasoning chains on the NQ dataset, with answers generated by Qwen3-8B. ``Vanilla" refers to no modification; ``Basic-All" denotes inserting four factual statements; ``Wrong-Few" means that one of them is replaced with a counterfactual statement; and ``Wrong-All" indicates that all statements are replaced with counterfactual ones.}
  \label{fig:factual_injection}
\end{figure}


Figure~\ref{fig:factual_injection} shows the impact of disrupting the fluency of reasoning chains on model judgments, comparing Vanilla (/w Think) with Basic-All.
For almost all the models except Llama3-8B, the injection of factual content leads to a significant decline in pass rates.  This indicates that disrupting the fluency of reasoning makes LLM judges more likely to deem an answer incorrect, and also suggests that the natural fluency of reasoning is a key factor in misleading models into judging answers as correct.
Interestingly, in Section~\S~\ref{sec:detailed_analysis} we find that weak judges are more easily misled by reasoning chains, while strong models exhibit more selective behavior. However, in this setting, when the fluency of the reasoning chain is clearly disrupted, all models become more likely to judge the answer as incorrect. This suggests that both strong and weak models are capable of identifying issues in the reasoning chain during evaluation, depending on how salient those issues are.

\subsubsection{Effects of Factuality}
Figure~\ref{fig:factual_injection} shows that, for almost all models, introducing counterfactual information reduces the pass rate, making judges more likely to deem answers incorrect. Due to space constraints, we provide more detailed results in Table~\ref{tab:fine_grained_errors}.
The results further show that, although counterfactual content lowers the pass rate, increasing the number of erroneous statements does not produce a consistent trend across levels. This suggests that the model’s ability to detect errors in reasoning chains is not solely determined by the quantity of incorrect statements, but may also depend on factors such as the type or quality of the errors. We leave a more detailed investigation of these factors for future work.

\subsubsection{Effects of Position}

To investigate which parts of the reasoning chain models tend to focus on during judgment, we analyze how the insertion position affects model decisions. The results in Table~\ref{tab:synthesized_reasoning} reveal a position sensitivity in judge behavior, with factual injections introduced as prefixes exerting a consistently stronger impact than those inserted as suffixes. Across all synthesized settings, suffix injections lead to smaller reductions in pass rate than prefixes. For example, under the \textit{Basic-All} condition, Qwen3-8B's pass rate drops to 61.0\% when counterfactual content is inserted as a prefix, but remains substantially higher at 86.0\% when the same content is appended as a suffix.

A similar pattern is observed for GPT-4o, where the pass rate drops to 18.4\% under the prefix condition but remains at 39.2\% for the suffix condition, closer to the 48.6\% with natural reasoning chains. 
We think this behavior is similar to that of humans: if an error appears at the beginning of the reasoning process, one tends to judge the entire chain as incorrect. In contrast, if issues arise only at the end, they do not affect the earlier reasoning steps, and thus have a relatively smaller impact. \looseness=-1

\section{Conclusions}
In this paper, we study how reasoning traces affect the reliability of LLM-based judges. We find that while reasoning provides richer signals, it also introduces systematic biases: weak judges are easily swayed by the presence and fluency of reasoning, often overestimating incorrect answers, whereas strong judges use reasoning more selectively but remain vulnerable to high-quality yet misleading chains.
Further analysis shows that both fluency and factuality strongly influence judgment, indicating that current judges struggle to distinguish true evidential value from superficial features. Overall, our results highlight that reasoning is a double-edged sword for evaluation and call for more robust judge designs that can critically verify, rather than be persuaded by, reasoning processes.

\section{Discussions}

\paragraph{Should model judgments of answer correctness be influenced by the reasoning process?}
A key question is whether the reasoning chain should influence a model’s judgment of answer correctness. On one hand, the primary goal in this work is to assess whether the final answer is correct, rather than whether the reasoning process itself is valid. Under this objective, if the model strictly follows the instruction, the reasoning chain should not affect the judgment outcome.
On the other hand, the reasoning process may provide additional signals that help the model better determine answer correctness. When the correctness of the reasoning aligns with that of the final answer, the reasoning chain can serve as a useful indicator.
However, existing reasoning models are typically trained with a focus on final outcomes—i.e., whether the answer is correct—without explicit supervision of reasoning quality. As a result, inconsistencies may arise between the reasoning and the final answer, especially in cases where the answer is correct but the reasoning is flawed. 
Therefore, while correct reasoning often implies a correct answer, incorrect reasoning does not necessarily imply an incorrect answer. The most critical case arises when the reasoning is flawed but the answer is actually correct. Rejecting such answers solely due to incorrect reasoning would lead to erroneous judgments.

\paragraph{If a model cannot determine whether an answer is correct, can it reliably judge the correctness of the reasoning process?}
Reasoning chains involve both domain knowledge related to the final answer and the logical steps connecting intermediate conclusions. If a model cannot reliably determine whether the final answer is correct, it is unclear whether it can accurately assess the correctness of the reasoning process.
Evaluating a reasoning chain typically requires stronger capabilities than judging a final answer, as it involves verifying multiple intermediate steps, detecting subtle logical errors, and ensuring factual consistency throughout. Therefore, a model that struggles with answer correctness is unlikely to reliably evaluate reasoning quality.
This suggests that using such models as judges of reasoning quality may be inherently unreliable, as limitations in knowledge and reasoning ability affect both answer verification and reasoning evaluation.

\section*{Limitations}
While our study provides systematic insights into the impact of reasoning chains on LLM-based judgment, there are several limitations to consider. First, our investigation is confined to text-based benchmarks. The influence of reasoning in multimodal contexts (e.g., vision-language tasks) remains unexplored and represents a promising direction for future work. Second, due to computational constraints, we did not prompt the judge models to generate their own reasoning chains prior to delivering a verdict. Instead, we restricted the models to providing direct judgments without intermediate reasoning steps. Finally, although we evaluated a diverse set of representative models, the rapid evolution of proprietary LLMs means our coverage is inevitably not exhaustive. Future studies could extend our findings to a broader array of emerging reasoning models.

\section*{Ethical Considerations}
All models and datasets used in this study are publicly available or accessed via official APIs. The datasets employed (NQ, HotpotQA, and GSM8K) are standard in the field and contain no personally identifiable information (PII) or offensive content. As this work focuses on evaluating model capabilities using existing resources, it introduces no additional societal risks or ethical concerns.

\section*{Acknowledgements}
This work was funded by the National Natural Science Foundation of China (NSFC) under Grant No. 62302486 and the Innovation Project of ICT CAS under Grant No. E361140.

\bibliography{custom}

@inproceedings{ni2024llms,
  title={When do llms need retrieval augmentation? mitigating llms’ overconfidence helps retrieval augmentation},
  author={Ni, Shiyu and Bi, Keping and Guo, Jiafeng and Cheng, Xueqi},
  booktitle={Findings of the Association for Computational Linguistics: ACL 2024},
  pages={11375--11388},
  year={2024}
}

@article{hendrycks2021measuring,
  title={Measuring mathematical problem solving with the math dataset},
  author={Hendrycks, Dan and Burns, Collin and Kadavath, Saurav and Arora, Akul and Basart, Steven and Tang, Eric and Song, Dawn and Steinhardt, Jacob},
  journal={arXiv preprint arXiv:2103.03874},
  year={2021}
}

@article{chen2025safer,
  title={Safer or Luckier? LLMs as Safety Evaluators Are Not Robust to Artifacts},
  author={Chen, Hongyu and Goldfarb-Tarrant, Seraphina},
  journal={arXiv preprint arXiv:2503.09347},
  year={2025}
}

@article{marioriyad2025silent,
  title={The Silent Judge: Unacknowledged Shortcut Bias in LLM-as-a-Judge},
  author={Marioriyad, Arash and Rohban, Mohammad Hossein and Baghshah, Mahdieh Soleymani},
  journal={arXiv preprint arXiv:2509.26072},
  year={2025}
}

@inproceedings{chiang-lee-2023-large,
    title = "Can Large Language Models Be an Alternative to Human Evaluations?",
    author = "Chiang, Cheng-Han  and
      Lee, Hung-yi",
    editor = "Rogers, Anna  and
      Boyd-Graber, Jordan  and
      Okazaki, Naoaki",
    booktitle = "Proceedings of the 61st Annual Meeting of the Association for Computational Linguistics (Volume 1: Long Papers)",
    month = jul,
    year = "2023",
    address = "Toronto, Canada",
    publisher = "Association for Computational Linguistics",
    url = "https://aclanthology.org/2023.acl-long.870/",
    doi = "10.18653/v1/2023.acl-long.870",
    pages = "15607--15631"
}

@inproceedings{liu-etal-2023-g,
    title = "{G}-Eval: {NLG} Evaluation using Gpt-4 with Better Human Alignment",
    author = "Liu, Yang  and
      Iter, Dan  and
      Xu, Yichong  and
      Wang, Shuohang  and
      Xu, Ruochen  and
      Zhu, Chenguang",
    editor = "Bouamor, Houda  and
      Pino, Juan  and
      Bali, Kalika",
    booktitle = "Proceedings of the 2023 Conference on Empirical Methods in Natural Language Processing",
    month = dec,
    year = "2023",
    address = "Singapore",
    publisher = "Association for Computational Linguistics",
    url = "https://aclanthology.org/2023.emnlp-main.153/",
    doi = "10.18653/v1/2023.emnlp-main.153",
    pages = "2511--2522"
}

@inproceedings{zheng2023judging,
    title={Judging {LLM}-as-a-Judge with {MT}-Bench and Chatbot Arena},
    author={Lianmin Zheng and Wei-Lin Chiang and Ying Sheng and Siyuan Zhuang and Zhanghao Wu and Yonghao Zhuang and Zi Lin and Zhuohan Li and Dacheng Li and Eric Xing and Hao Zhang and Joseph E. Gonzalez and Ion Stoica},
    booktitle={Thirty-seventh Conference on Neural Information Processing Systems Datasets and Benchmarks Track},
    year={2023},
    url={https://openreview.net/forum?id=uccHPGDlao}
}

@inproceedings{dubois2023alpacafarm,
    title={AlpacaFarm: A Simulation Framework for Methods that Learn from Human Feedback},
    author={Yann Dubois and Xuechen Li and Rohan Taori and Tianyi Zhang and Ishaan Gulrajani and Jimmy Ba and Carlos Guestrin and Percy Liang and Tatsunori Hashimoto},
    booktitle={Thirty-seventh Conference on Neural Information Processing Systems},
    year={2023},
    url={https://openreview.net/forum?id=4hturzLcKX}
}

@article{wei2022chain,
  title={Chain-of-thought prompting elicits reasoning in large language models},
  author={Wei, Jason and Wang, Xuezhi and Schuurmans, Dale and Bosma, Maarten and Xia, Fei and Chi, Ed and Le, Quoc V and Zhou, Denny and others},
  journal={Advances in neural information processing systems},
  volume={35},
  pages={24824--24837},
  year={2022}
}

@article{zhang2025reasoning,
  title={Reasoning Models Know When They're Right: Probing Hidden States for Self-Verification},
  author={Zhang, Anqi and Chen, Yulin and Pan, Jane and Zhao, Chen and Panda, Aurojit and Li, Jinyang and He, He},
  journal={arXiv preprint arXiv:2504.05419},
  year={2025}
}

@article{openai2024gpt4ocard,
  title={Gpt-4o system card},
  author={Hurst, Aaron and Lerer, Adam and Goucher, Adam P and Perelman, Adam and Ramesh, Aditya and Clark, Aidan and Ostrow, AJ and Welihinda, Akila and Hayes, Alan and Radford, Alec and others},
  journal={arXiv preprint arXiv:2410.21276},
  year={2024}
}

@techreport{meta_llama31_modelcard,
  title        = {Llama 3.1 Model Card},
  author       = {{AI@Meta}},
  institution  = {Meta AI},
  year         = {2024},
  url          = {https://github.com/meta-llama/llama-models/blob/main/models/llama3_1/MODEL_CARD.md},
  note         = {Online documentation, accessed 2026-01-05}
}

@article{yang2025qwen3,
  title={Qwen3 technical report},
  author={Yang, An and Li, Anfeng and Yang, Baosong and Zhang, Beichen and Hui, Binyuan and Zheng, Bo and Yu, Bowen and Gao, Chang and Huang, Chengen and Lv, Chenxu and others},
  journal={arXiv preprint arXiv:2505.09388},
  year={2025}
}

@misc{qwen2025qwen25technicalreport,
  title={Qwen2.5 Technical Report}, 
  author={Qwen and : and An Yang and Baosong Yang and Beichen Zhang and Binyuan Hui and Bo Zheng and Bowen Yu and Chengyuan Li and Dayiheng Liu and Fei Huang and Haoran Wei and Huan Lin and Jian Yang and Jianhong Tu and Jianwei Zhang and Jianxin Yang and Jiaxi Yang and Jingren Zhou and Junyang Lin and Kai Dang and Keming Lu and Keqin Bao and Kexin Yang and Le Yu and Mei Li and Mingfeng Xue and Pei Zhang and Qin Zhu and Rui Men and Runji Lin and Tianhao Li and Tianyi Tang and Tingyu Xia and Xingzhang Ren and Xuancheng Ren and Yang Fan and Yang Su and Yichang Zhang and Yu Wan and Yuqiong Liu and Zeyu Cui and Zhenru Zhang and Zihan Qiu},
  year={2025},
  eprint={2412.15115},
  archivePrefix={arXiv},
  primaryClass={cs.CL},
  url={https://arxiv.org/abs/2412.15115}, 
}

@misc{deepseekai2025deepseekv3technicalreport,
  title={DeepSeek-V3 Technical Report}, 
  author={DeepSeek-AI and Aixin Liu and Bei Feng and Bing Xue and Bingxuan Wang and Bochao Wu and Chengda Lu and Chenggang Zhao and Chengqi Deng and Chenyu Zhang and Chong Ruan and Damai Dai and Daya Guo and Dejian Yang and Deli Chen and Dongjie Ji and Erhang Li and Fangyun Lin and Fucong Dai and Fuli Luo and Guangbo Hao and Guanting Chen and Guowei Li and H. Zhang and Han Bao and Hanwei Xu and Haocheng Wang and Haowei Zhang and Honghui Ding and Huajian Xin and Huazuo Gao and Hui Li and Hui Qu and J. L. Cai and Jian Liang and Jianzhong Guo and Jiaqi Ni and Jiashi Li and Jiawei Wang and Jin Chen and Jingchang Chen and Jingyang Yuan and Junjie Qiu and Junlong Li and Junxiao Song and Kai Dong and Kai Hu and Kaige Gao and Kang Guan and Kexin Huang and Kuai Yu and Lean Wang and Lecong Zhang and Lei Xu and Leyi Xia and Liang Zhao and Litong Wang and Liyue Zhang and Meng Li and Miaojun Wang and Mingchuan Zhang and Minghua Zhang and Minghui Tang and Mingming Li and Ning Tian and Panpan Huang and Peiyi Wang and Peng Zhang and Qiancheng Wang and Qihao Zhu and Qinyu Chen and Qiushi Du and R. J. Chen and R. L. Jin and Ruiqi Ge and Ruisong Zhang and Ruizhe Pan and Runji Wang and Runxin Xu and Ruoyu Zhang and Ruyi Chen and S. S. Li and Shanghao Lu and Shangyan Zhou and Shanhuang Chen and Shaoqing Wu and Shengfeng Ye and Shengfeng Ye and Shirong Ma and Shiyu Wang and Shuang Zhou and Shuiping Yu and Shunfeng Zhou and Shuting Pan and T. Wang and Tao Yun and Tian Pei and Tianyu Sun and W. L. Xiao and Wangding Zeng and Wanjia Zhao and Wei An and Wen Liu and Wenfeng Liang and Wenjun Gao and Wenqin Yu and Wentao Zhang and X. Q. Li and Xiangyue Jin and Xianzu Wang and Xiao Bi and Xiaodong Liu and Xiaohan Wang and Xiaojin Shen and Xiaokang Chen and Xiaokang Zhang and Xiaosha Chen and Xiaotao Nie and Xiaowen Sun and Xiaoxiang Wang and Xin Cheng and Xin Liu and Xin Xie and Xingchao Liu and Xingkai Yu and Xinnan Song and Xinxia Shan and Xinyi Zhou and Xinyu Yang and Xinyuan Li and Xuecheng Su and Xuheng Lin and Y. K. Li and Y. Q. Wang and Y. X. Wei and Y. X. Zhu and Yang Zhang and Yanhong Xu and Yanhong Xu and Yanping Huang and Yao Li and Yao Zhao and Yaofeng Sun and Yaohui Li and Yaohui Wang and Yi Yu and Yi Zheng and Yichao Zhang and Yifan Shi and Yiliang Xiong and Ying He and Ying Tang and Yishi Piao and Yisong Wang and Yixuan Tan and Yiyang Ma and Yiyuan Liu and Yongqiang Guo and Yu Wu and Yuan Ou and Yuchen Zhu and Yuduan Wang and Yue Gong and Yuheng Zou and Yujia He and Yukun Zha and Yunfan Xiong and Yunxian Ma and Yuting Yan and Yuxiang Luo and Yuxiang You and Yuxuan Liu and Yuyang Zhou and Z. F. Wu and Z. Z. Ren and Zehui Ren and Zhangli Sha and Zhe Fu and Zhean Xu and Zhen Huang and Zhen Zhang and Zhenda Xie and Zhengyan Zhang and Zhewen Hao and Zhibin Gou and Zhicheng Ma and Zhigang Yan and Zhihong Shao and Zhipeng Xu and Zhiyu Wu and Zhongyu Zhang and Zhuoshu Li and Zihui Gu and Zijia Zhu and Zijun Liu and Zilin Li and Ziwei Xie and Ziyang Song and Ziyi Gao and Zizheng Pan},
  year={2025},
  eprint={2412.19437},
  archivePrefix={arXiv},
  primaryClass={cs.CL},
  url={https://arxiv.org/abs/2412.19437}, 
}

@misc{zai_glm4_32b,
  author       = {{Zhipu AI}},
  title        = {GLM-4-32B Model Card},
  year         = {2025},
  howpublished = {\url{https://huggingface.co/zai-org/GLM-4-32B-0414}},
}

@techreport{anthropic_claude_opus_4_5_2025,
  title        = {Claude Opus 4.5 System Card},
  author       = {{Anthropic}},
  institution  = {Anthropic},
  year         = {2025},
  month        = {March},
  url          = {https://assets.anthropic.com/m/64823ba7485345a7/Claude-Opus-4-5-System-Card.pdf},
  note         = {Accessed: 2026-01-05}
}

@article{kwiatkowski2019natural,
  title={Natural questions: a benchmark for question answering research},
  author={Kwiatkowski, Tom and Palomaki, Jennimaria and Redfield, Olivia and Collins, Michael and Parikh, Ankur and Alberti, Chris and Epstein, Danielle and Polosukhin, Illia and Devlin, Jacob and Lee, Kenton and others},
  journal={Transactions of the Association for Computational Linguistics},
  volume={7},
  pages={453--466},
  year={2019},
  publisher={MIT Press One Rogers Street, Cambridge, MA 02142-1209, USA journals-info~…}
}

@inproceedings{yang2018hotpotqa,
  title={HotpotQA: A dataset for diverse, explainable multi-hop question answering},
  author={Yang, Zhilin and Qi, Peng and Zhang, Saizheng and Bengio, Yoshua and Cohen, William and Salakhutdinov, Ruslan and Manning, Christopher D},
  booktitle={Proceedings of the 2018 conference on empirical methods in natural language processing},
  pages={2369--2380},
  year={2018}
}

@article{cobbe2021training,
  title={Training verifiers to solve math word problems},
  author={Cobbe, Karl and Kosaraju, Vineet and Bavarian, Mohammad and Chen, Mark and Jun, Heewoo and Kaiser, Lukasz and Plappert, Matthias and Tworek, Jerry and Hilton, Jacob and Nakano, Reiichiro and others},
  journal={arXiv preprint arXiv:2110.14168},
  year={2021}
}

@inproceedings{manas2024improving,
  title={Improving automatic vqa evaluation using large language models},
  author={Ma{\~n}as, Oscar and Krojer, Benno and Agrawal, Aishwarya},
  booktitle={Proceedings of the AAAI Conference on Artificial Intelligence},
  volume={38},
  pages={4171--4179},
  year={2024}
}

@article{brown2020language,
  title={Language models are few-shot learners},
  author={Brown, Tom and Mann, Benjamin and Ryder, Nick and Subbiah, Melanie and Kaplan, Jared D and Dhariwal, Prafulla and Neelakantan, Arvind and Shyam, Pranav and Sastry, Girish and Askell, Amanda and others},
  journal={Advances in neural information processing systems},
  volume={33},
  pages={1877--1901},
  year={2020}
}

@article{verga2024replacing,
  title={Replacing judges with juries: Evaluating llm generations with a panel of diverse models},
  author={Verga, Pat and Hofstatter, Sebastian and Althammer, Sophia and Su, Yixuan and Piktus, Aleksandra and Arkhangorodsky, Arkady and Xu, Minjie and White, Naomi and Lewis, Patrick},
  journal={arXiv preprint arXiv:2404.18796},
  year={2024}
}

@inproceedings{huang-etal-2024-chatgpt,
  title={ChatGPT rates natural language explanation quality like humans: But on which scales?},
  author={Huang, Fan and Kwak, Haewoon and Park, Kunwoo and An, Jisun},
  booktitle={Proceedings of the 2024 Joint International Conference on Computational Linguistics, Language Resources and Evaluation (LREC-COLING 2024)},
  pages={3111--3132},
  year={2024}
}

@inproceedings{ni2025towards,
  title={Towards fully exploiting llm internal states to enhance knowledge boundary perception},
  author={Ni, Shiyu and Bi, Keping and Guo, Jiafeng and Yu, Lulu and Bi, Baolong and Cheng, Xueqi},
  booktitle={Proceedings of the 63rd Annual Meeting of the Association for Computational Linguistics (Volume 1: Long Papers)},
  pages={24315--24329},
  year={2025}
}

@article{shi2025deep,
  title={Deep research: A systematic survey},
  author={Shi, Zhengliang and Chen, Yiqun and Li, Haitao and Sun, Weiwei and Ni, Shiyu and Lyu, Yougang and Fan, Run-Ze and Jin, Bowen and Weng, Yixuan and Zhu, Minjun and others},
  journal={arXiv preprint arXiv:2512.02038},
  year={2025}
}

@inproceedings{pavlovic-poesio-2024-effectiveness,
  title={The effectiveness of LLMs as annotators: A comparative overview and empirical analysis of direct representation},
  author={Pavlovic, Maja and Poesio, Massimo},
  booktitle={Proceedings of the 3rd Workshop on Perspectivist Approaches to NLP (NLPerspectives)@ LREC-COLING 2024},
  pages={100--110},
  year={2024}
}

@inproceedings{tan2025judgebench,
    title={JudgeBench: A Benchmark for Evaluating {LLM}-Based Judges},
    author={Sijun Tan and Siyuan Zhuang and Kyle Montgomery and William Yuan Tang and Alejandro Cuadron and Chenguang Wang and Raluca Popa and Ion Stoica},
    booktitle={The Thirteenth International Conference on Learning Representations},
    year={2025},
    url={https://openreview.net/forum?id=G0dksFayVq}
}

@inproceedings{wang2024large,
  title={Large language models are not fair evaluators},
  author={Wang, Peiyi and Li, Lei and Chen, Liang and Cai, Zefan and Zhu, Dawei and Lin, Binghuai and Cao, Yunbo and Kong, Lingpeng and Liu, Qi and Liu, Tianyu and others},
  booktitle={Proceedings of the 62nd Annual Meeting of the Association for Computational Linguistics (Volume 1: Long Papers)},
  pages={9440--9450},
  year={2024}
}

@article{wang2023pandalm,
  title={Pandalm: An automatic evaluation benchmark for llm instruction tuning optimization},
  author={Wang, Yidong and Yu, Zhuohao and Zeng, Zhengran and Yang, Linyi and Wang, Cunxiang and Chen, Hao and Jiang, Chaoya and Xie, Rui and Wang, Jindong and Xie, Xing and others},
  journal={arXiv preprint arXiv:2306.05087},
  year={2023}
}

@inproceedings{kim2023prometheus,
  title={Prometheus: Inducing fine-grained evaluation capability in language models},
  author={Kim, Seungone and Shin, Jamin and Cho, Yejin and Jang, Joel and Longpre, Shayne and Lee, Hwaran and Yun, Sangdoo and Shin, Seongjin and Kim, Sungdong and Thorne, James and others},
  booktitle={The Twelfth International Conference on Learning Representations},
  year={2023}
}

@article{zhu2023judgelm,
  title={Judgelm: Fine-tuned large language models are scalable judges},
  author={Zhu, Lianghui and Wang, Xinggang and Wang, Xinlong},
  journal={arXiv preprint arXiv:2310.17631},
  year={2023}
}

@article{li2023generative,
  title={Generative judge for evaluating alignment},
  author={Li, Junlong and Sun, Shichao and Yuan, Weizhe and Fan, Run-Ze and Zhao, Hai and Liu, Pengfei},
  journal={arXiv preprint arXiv:2310.05470},
  year={2023}
}

@inproceedings{du2024improving,
  title={Improving factuality and reasoning in language models through multiagent debate},
  author={Du, Yilun and Li, Shuang and Torralba, Antonio and Tenenbaum, Joshua B and Mordatch, Igor},
  booktitle={Forty-first international conference on machine learning},
  year={2024}
}

@article{chan2023chateval,
  title={Chateval: Towards better llm-based evaluators through multi-agent debate},
  author={Chan, Chi-Min and Chen, Weize and Su, Yusheng and Yu, Jianxuan and Xue, Wei and Zhang, Shanghang and Fu, Jie and Liu, Zhiyuan},
  journal={arXiv preprint arXiv:2308.07201},
  year={2023}
}

@article{madaan2023self,
  title={Self-refine: Iterative refinement with self-feedback},
  author={Madaan, Aman and Tandon, Niket and Gupta, Prakhar and Hallinan, Skyler and Gao, Luyu and Wiegreffe, Sarah and Alon, Uri and Dziri, Nouha and Prabhumoye, Shrimai and Yang, Yiming and others},
  journal={Advances in neural information processing systems},
  volume={36},
  pages={46534--46594},
  year={2023}
}

@article{kojima2022large,
  title={Large language models are zero-shot reasoners},
  author={Kojima, Takeshi and Gu, Shixiang Shane and Reid, Machel and Matsuo, Yutaka and Iwasawa, Yusuke},
  journal={Advances in neural information processing systems},
  volume={35},
  pages={22199--22213},
  year={2022}
}

@article{wang2022self,
  title={Self-consistency improves chain of thought reasoning in language models},
  author={Wang, Xuezhi and Wei, Jason and Schuurmans, Dale and Le, Quoc and Chi, Ed and Narang, Sharan and Chowdhery, Aakanksha and Zhou, Denny},
  journal={arXiv preprint arXiv:2203.11171},
  year={2022}
}

@article{zhou2022least,
  title={Least-to-most prompting enables complex reasoning in large language models},
  author={Zhou, Denny and Sch{\"a}rli, Nathanael and Hou, Le and Wei, Jason and Scales, Nathan and Wang, Xuezhi and Schuurmans, Dale and Cui, Claire and Bousquet, Olivier and Le, Quoc and others},
  journal={arXiv preprint arXiv:2205.10625},
  year={2022}
}

@inproceedings{press2023measuring,
  title={Measuring and narrowing the compositionality gap in language models},
  author={Press, Ofir and Zhang, Muru and Min, Sewon and Schmidt, Ludwig and Smith, Noah A and Lewis, Mike},
  booktitle={Findings of the Association for Computational Linguistics: EMNLP 2023},
  pages={5687--5711},
  year={2023}
}

@article{zelikman2022star,
  title={Star: Bootstrapping reasoning with reasoning},
  author={Zelikman, Eric and Wu, Yuhuai and Mu, Jesse and Goodman, Noah},
  journal={Advances in Neural Information Processing Systems},
  volume={35},
  pages={15476--15488},
  year={2022}
}

@article{yao2023tree,
  title={Tree of thoughts: Deliberate problem solving with large language models},
  author={Yao, Shunyu and Yu, Dian and Zhao, Jeffrey and Shafran, Izhak and Griffiths, Tom and Cao, Yuan and Narasimhan, Karthik},
  journal={Advances in neural information processing systems},
  volume={36},
  pages={11809--11822},
  year={2023}
}

@inproceedings{besta2024graph,
  title={Graph of thoughts: Solving elaborate problems with large language models},
  author={Besta, Maciej and Blach, Nils and Kubicek, Ales and Gerstenberger, Robert and Podstawski, Michal and Gianinazzi, Lukas and Gajda, Joanna and Lehmann, Tomasz and Niewiadomski, Hubert and Nyczyk, Piotr and others},
  booktitle={Proceedings of the AAAI conference on artificial intelligence},
  volume={38},
  pages={17682--17690},
  year={2024}
}

@article{shinn2023reflexion,
  title={Reflexion: Language agents with verbal reinforcement learning},
  author={Shinn, Noah and Cassano, Federico and Gopinath, Ashwin and Narasimhan, Karthik and Yao, Shunyu},
  journal={Advances in neural information processing systems},
  volume={36},
  pages={8634--8652},
  year={2023}
}

@inproceedings{chiang2024chatbot,
  title={Chatbot arena: An open platform for evaluating llms by human preference},
  author={Chiang, Wei-Lin and Zheng, Lianmin and Sheng, Ying and Angelopoulos, Anastasios Nikolas and Li, Tianle and Li, Dacheng and Zhu, Banghua and Zhang, Hao and Jordan, Michael and Gonzalez, Joseph E and others},
  booktitle={Forty-first International Conference on Machine Learning},
  year={2024}
}

\appendix
\section{Prompts}
\label{sec:prompts}

\begin{promptbox}{short\_qa}
\raggedright
Answer the following question based on your internal knowledge with one or few words.
\\[1em]
Question:\{question\}
\end{promptbox}

\begin{promptbox}{llm\_judge\_without\_think}
\raggedright
Judge whether the following answer about the question is correct. If you are sure the answer is correct, say \texttt{certain}. If not, please say \texttt{uncertain}. Just give your judgment without any other words.
\\[1em]
Question:\{question\}
\\[1em]
Answer:\{prediction\}
\end{promptbox}

\begin{promptbox}{llm\_judge\_with\_think}
\raggedright
Judge whether the following answer about the question is correct. The content inside <think></think> represents the reasoning process of the model, while the content after </think> is the answer provided by the model. If you are sure the answer is correct, say \texttt{certain}. If not, please say \texttt{uncertain}. Just give your judgment without any other words.
\\[1em]
Question:\{question\}
\\[1em]
Answer:\{prediction\}
\end{promptbox}

\begin{promptbox}{llm\_selfjudge}
\raggedright
Answer the following question based on your internal knowledge with one or few words. Then, judge whether your answer is correct. If you are sure the answer is correct, say \texttt{certain}. If not, please say \texttt{uncertain}. Your output should be in the following format: \\
Answer: <your answer> \\ 
Judge: <certain/uncertain>
\\[1em]
Question:\{question\}
\end{promptbox}

\begin{promptbox}{basic\_all}
\raggedright
The Earth orbits the Sun once every 365 days, producing the cycle of the seasons. Water freezes at 0 degrees Celsius and boils at 100 degrees Celsius at standard atmospheric pressure. Humans typically have 206 bones in the adult skeleton. The Pacific Ocean is the largest ocean on Earth, and Mount Everest is the tallest mountain above sea level.
\end{promptbox}

\begin{promptbox}{wrong\_few / wrong\_1}
\raggedright
The Earth orbits the Sun once every \textcolor{red}{100} days, producing the cycle of the seasons. Water freezes at 0 degrees Celsius and boils at 100 degrees Celsius at standard atmospheric pressure. Humans typically have 206 bones in the adult skeleton. The Pacific Ocean is the largest ocean on Earth, and Mount Everest is the tallest mountain above sea level.
\end{promptbox}

\begin{promptbox}{wrong\_2}
\raggedright
The Earth orbits the Sun once every \textcolor{red}{100} days, producing the cycle of the seasons. Water freezes at \textcolor{red}{10} degrees Celsius and boils at 100 degrees Celsius at standard atmospheric pressure. Humans typically have 206 bones in the adult skeleton. The Pacific Ocean is the largest ocean on Earth, and Mount Everest is the tallest mountain above sea level.
\end{promptbox}

\begin{promptbox}{wrong\_3}
\raggedright
The Earth orbits the Sun once every \textcolor{red}{100} days, producing the cycle of the seasons. Water freezes at \textcolor{red}{10} degrees Celsius and boils at 100 degrees Celsius at standard atmospheric pressure. Humans typically have \textcolor{red}{100} bones in the adult skeleton. The Pacific Ocean is the \textcolor{red}{smallest} ocean on Earth, and Mount Everest is the tallest mountain above sea level.
\end{promptbox}

\begin{promptbox}{wrong\_all}
\raggedright
The Earth orbits the Sun once every \textcolor{red}{100} days, producing the cycle of the seasons. Water freezes at \textcolor{red}{10} degrees Celsius and boils at 100 degrees Celsius at standard atmospheric pressure. Humans typically have \textcolor{red}{100} bones in the adult skeleton. The Pacific Ocean is the largest ocean on Earth, and Mount Everest is the tallest mountain above sea level.
\end{promptbox}

\section{Results Using Other Generators}
\label{sec:Other_Generator}
Table~\ref{tab:Qwen3-14B_judge} to Table~\ref{tab:DeepSeek-v3.1_judge} show results on NQ, HotpotQA, and GSM8K using different generators,including Qwen3-14B, Qwen3-32B, and DeepSeek-v3.1, respectively.

Table ~\ref{tab:math500_results} shows results on MATH500 using Qwen3-8B as the generator, and Table~\ref{tab:math500_balanced_results} further evaluates model behavior on a fully balanced subset of MATH500, enabling a more controlled analysis of potential biases.

Table ~\ref{tab:fine_grained_errors} provides a fine-grained analysis of judge pass rates on NQ under progressively injected erroneous sentences in reasoning chains (Basic-All, Wrong-1 to Wrong-All).

\begin{figure*}[t]
  \centering
  \includegraphics[width=\textwidth]{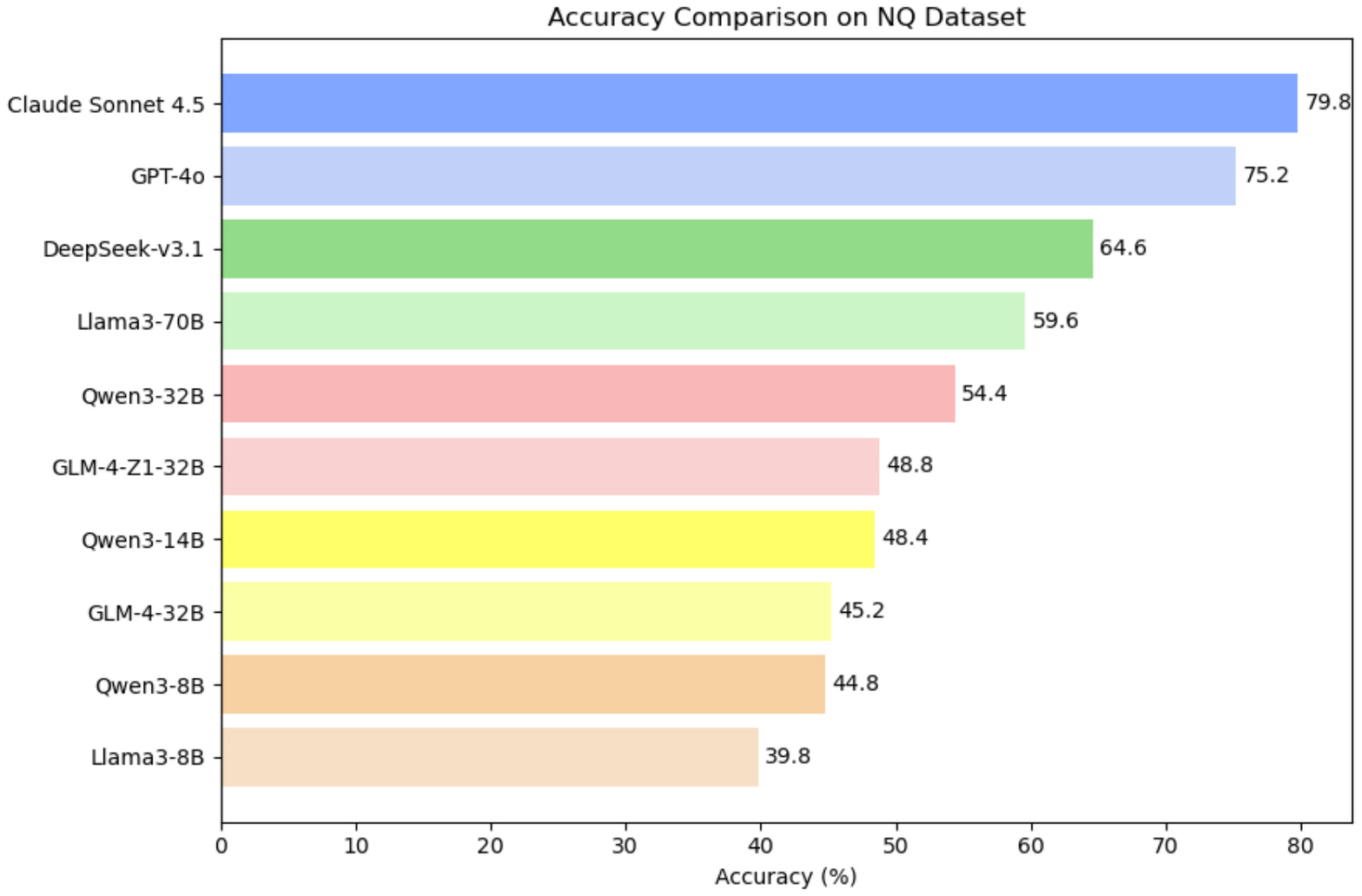}
  \caption{Accuracy (\%) comparison of different models on NQ dataset for question answering, evaluated on 500 filtered samples with verified answers.}
  \label{fig:Acc_Compare}
\end{figure*}

Figure ~\ref{fig:Acc_Compare} shows the question answering accuracy of different models as answer generators on NQ, evaluated on a filtered subset of 500 samples with verified answers. Based on these results, we categorize Claude Sonnet 4.5, GPT-4o, and DeepSeek-v3.1 as strong models, and the remaining models as weak models.

\begin{table*}[t]
\centering
\caption{Evaluation results(\%) of LLM-as-a-Judge behavior with and without reasoning chains across 
factual and mathematical datasets, with all answers generated by Qwen3-14B.}
\scriptsize

\setlength{\tabcolsep}{6pt}
\begin{tabular}{l c l cc cc cc cc}
\toprule
\multirow{2}{*}{Dataset}
& \multirow{2}{*}{Acc}
& \multirow{2}{*}{Judge Models}
& \multicolumn{2}{c}{Alignment}
& \multicolumn{2}{c}{Pass Rate}
& \multicolumn{2}{c}{Overconfidence}
& \multicolumn{2}{c}{Conservativeness} \\
\cmidrule(lr){4-5} \cmidrule(lr){6-7} \cmidrule(lr){8-9} \cmidrule(lr){10-11}
 &  & 
& w/o Think & w/ Think
& w/o Think & w/ Think
& w/o Think & w/ Think
& w/o Think & w/ Think \\
\midrule

\multirow{10}{*}{NQ}
& \multirow{10}{*}{31.2}
& Qwen3-8B
& 57.8 & 39.2 & 59.8 & 90.8 & 35.4 & 60.2 & 6.8 & 0.6 \\

& & Qwen3-14B
& 55.2 & 40.6 & 67.2 & 89.8 & 40.4 & 59.0 & 4.4 & 0.4 \\

& & Qwen3-32B
& 40.8 & 41.2 & 88.0 & 89.6 & 58.0 & 58.6 & 1.2 & 0.2 \\

& & Llama3-8B
& 40.2 & 32.6 & 87.8 & 98.6 & 58.2 & 67.4 & 1.6 & 0.0 \\

& & Llama3-70B
& 51.0 & 41.4 & 76.6 & 87.0 & 47.2 & 57.2 & 1.8 & 1.4 \\

& & GLM4-32B
& 47.6 & 37.0 & 82.0 & 94.2 & 51.6 & 63.0 & 0.8 & 0.0 \\

& & GLM4-Z1-32B
& 75.2 & 50.8 & 36.8 & 72.4 & 15.2 & 45.2 & 9.6 & 4.0 \\

\cmidrule(lr){3-11}

& & GPT-4o
& 73.6 & 70.2 & 45.2 & 51.8 & 20.2 & 25.2 & 6.2 & 4.6 \\

& & DeepSeek-v3.1
& 62.8 & 61.8 & 61.6 & 62.2 & 33.8 & 34.6 & 3.4 & 3.6 \\

& & Claude Sonnet 4.5
& 77.2 & 77.2 & 14.0 & 10.4 & 2.8 & 10.0 & 20.0 & 21.8 \\

\midrule
\multirow{10}{*}{HotpotQA}
& \multirow{10}{*}{30.8}
& Qwen3-8B
& 56.2 & 47.0 & 57.4 & 75.0 & 35.2 & 48.6 & 8.6 & 4.4 \\

& & Qwen3-14B
& 53.6 & 49.0 & 66.0 & 75.8 & 40.8 & 48.0 & 5.6 & 3.0 \\

& & Qwen3-32B
& 44.0 & 49.4 & 83.2 & 75.0 & 54.2 & 47.4 & 1.8 & 3.2 \\

& & Llama3-8B
& 44.6 & 38.8 & 77.0 & 85.6 & 50.8 & 58.0 & 4.6 & 3.2 \\

& & Llama3-70B
& 52.6 & 49.6 & 69.4 & 76.0 & 43.0 & 47.8 & 4.4 & 2.6 \\

& & GLM4-32B
& 50.2 & 46.2 & 75.8 & 82.2 & 47.4 & 52.6 & 2.4 & 1.2 \\

& & GLM4-Z1-32B
& 74.6 & 59.2 & 14.6 & 50.8 & 4.6 & 30.4 & 20.8 & 10.4 \\

\cmidrule(lr){3-11}

& & GPT-4o
& 77.2 & 75.6 & 29.2 & 32.4 & 10.6 & 13.0 & 12.2 & 11.4 \\

& & DeepSeek
& 62.4 & 64.6 & 55.6 & 51.4 & 31.2 & 28.0 & 6.4 & 7.4 \\

& & Claude
& 75.0 & 70.8 & 9.8 & 3.6 & 2.0 & 1.0 & 23.0 & 28.2 \\

\midrule
\multirow{10}{*}{GSM8K}
& \multirow{10}{*}{94.0}
& Qwen3-8B
& 72.8 & 94.8 & 72.4 & 99.2 & 2.8 & 5.2 & 24.4 & 0.0 \\

& & Qwen3-14B
& 70.4 & 94.8 & 71.2 & 99.2 & 3.4 & 5.2 & 26.2 & 0.0 \\

& & Qwen3-32B
& 86.8 & 94.6 & 89.6 & 99.4 & 4.4 & 5.4 & 8.8 & 0.0 \\

& & Llama3-8B
& 89.8 & 94.8 & 93.0 & 98.8 & 4.6 & 5.0 & 5.6 & 0.2 \\

& & Llama3-70B
& 83.0 & 95.2 & 85.8 & 98.0 & 4.4 & 4.4 & 12.6 & 0.4 \\

& & GLM4-32B
& 89.2 & 95.0 & 92.8 & 99.0 & 4.8 & 5.0 & 6.0 & 0.0 \\

& & GLM4-Z1-32B
& 47.4 & 86.4 & 45.8 & 87.6 & 2.2 & 3.6 & 50.4 & 10.0 \\

\cmidrule(lr){3-11}

& & GPT-4o
& 93.4 & 92.0 & 94.2 & 94.0 & 3.4 & 4.0 & 3.2 & 4.0 \\

& & DeepSeek-v3.1
& 94.8 & 94.8 & 98.4 & 98.4 & 4.8 & 4.8 & 0.4 & 0.4 \\

& & Claude Sonnet 4.5
& 50.0 & 50.0 & 46.0 & 46.0 & 1.0 & 1.0 & 49.0 & 49.0 \\

\bottomrule
\end{tabular}

\label{tab:Qwen3-14B_judge}
\end{table*}

\begin{table*}[t]
\centering
\caption{Evaluation results(\%) of LLM-as-a-Judge behavior with and without reasoning chains across factual and mathematical datasets, with all answers generated by Qwen3-32B.}
\scriptsize

\setlength{\tabcolsep}{6pt}
\begin{tabular}{l c l cc cc cc cc}
\toprule
\multirow{2}{*}{Dataset}
& \multirow{2}{*}{Acc}
& \multirow{2}{*}{Judge Models}
& \multicolumn{2}{c}{Alignment}
& \multicolumn{2}{c}{Pass Rate}
& \multicolumn{2}{c}{Overconfidence}
& \multicolumn{2}{c}{Conservativeness} \\
\cmidrule(lr){4-5} \cmidrule(lr){6-7} \cmidrule(lr){8-9} \cmidrule(lr){10-11}
 &  & 
& w/o Think & w/ Think
& w/o Think & w/ Think
& w/o Think & w/ Think
& w/o Think & w/ Think \\
\midrule

\multirow{10}{*}{NQ}
& \multirow{10}{*}{35.6}
& Qwen3-8B
& 54.6 & 41.2 & 63.8 & 93.6 & 36.8 & 58.4 & 8.6 & 0.4 \\

& & Qwen3-14B
& 56.2 & 40.6 & 69.4 & 94.2 & 38.8 & 59.0 & 5.0 & 0.4 \\

& & Qwen3-32B
& 42.6 & 41.4 & 90.2 & 93.4 & 56.0 & 58.2 & 1.4 & 0.4 \\

& & Llama3-8B
& 39.8 & 36.8 & 92.2 & 98.4 & 58.4 & 63.0 & 1.8 & 0.2 \\

& & Llama3-70B
& 56.2 & 47.2 & 73.4 & 86.8 & 40.8 & 52.0 & 3.0 & 0.8 \\

& & GLM4-32B
& 46.8 & 41.6 & 86.0 & 92.8 & 51.8 & 57.8 & 1.4 & 0.6 \\

& & GLM4-Z1-32B
& 71.8 & 49.8 & 36.6 & 82.2 & 14.6 & 48.4 & 13.6 & 1.8 \\

\cmidrule(lr){3-11}

& & GPT-4o
& 74.6 & 72.0 & 51.8 & 56.8 & 20.8 & 24.6 & 4.6 & 3.4 \\

& & DeepSeek-v3.1
& 64.0 & 67.4 & 61.2 & 52.6 & 30.8 & 24.8 & 5.2 & 7.8 \\

& & Claude Sonnet 4.5
& 73.0 & 70.4 & 11.8 & 8.8 & 1.6 & 1.4 & 25.4 & 28.2 \\

\midrule
\multirow{10}{*}{HotpotQA}
& \multirow{10}{*}{34.4}
& Qwen3-8B
& 58.4 & 43.2 & 57.6 & 85.2 & 32.4 & 53.8 & 9.2 & 3.0 \\

& & Qwen3-14B
& 59.6 & 41.0 & 60.0 & 87.4 & 33.0 & 56.0 & 7.4 & 3.0 \\

& & Qwen3-32B
& 49.0 & 46.6 & 83.4 & 83.4 & 50.0 & 51.2 & 1.0 & 2.2 \\

& & Llama3-8B
& 47.6 & 39.2 & 78.8 & 92.0 & 48.4 & 59.2 & 4.0 & 1.6 \\

& & Llama3-70B
& 57.6 & 48.6 & 66.4 & 81.8 & 37.2 & 49.4 & 5.2 & 2.0 \\

& & GLM4-32B
& 54.8 & 45.6 & 74.8 & 87.6 & 42.8 & 53.8 & 2.4 & 0.6 \\

& & GLM4-Z1-32B
& --   & 54.0 & --   & 65.2 & --   & 38.4 & --   & 7.6 \\

\cmidrule(lr){3-11}

& & GPT-4o
& 78.0 & 76.2 & 30.0 & 36.6 & 8.8 & 13.0 & 13.2 & 10.8 \\

& & DeepSeek-v3.1
& 60.2 & 64.0 & 63.4 & 54.4 & 34.4 & 28.0 & 5.4 & 8.0 \\

& & Claude Sonnet 4.5
& 71.2 & 69.8 & 8.4 & 5.0 & 0.4 & 1.4 & 27.4 & 29.8 \\

\midrule
\multirow{10}{*}{GSM8K}
& \multirow{10}{*}{95.8}
& Qwen3-8B
& 81.6 & 95.6 & 81.8 & 99.8 & 2.2 & 4.2 & 16.2 & 0.2 \\

& & Qwen3-14B
& 84.4 & 95.6 & 85.0 & 99.8 & 2.4 & 4.2 & 13.2 & 0.2 \\

& & Qwen3-32B
& 90.4 & 96.0 & 93.8 & 99.8 & 3.8 & 4.0 & 5.8 & 0.0 \\

& & Llama3-8B
& 94.4 & 96.2 & 98.2 & 99.2 & 4.0 & 3.6 & 1.6 & 0.2 \\

& & Llama3-70B
& 87.0 & 95.6 & 90.0 & 99.4 & 3.6 & 4.0 & 9.4 & 0.4 \\

& & GLM4-32B
& 94.2 & 96.0 & 98.4 & 99.8 & 4.2 & 4.0 & 1.6 & 0.0 \\

& & GLM4-Z1-32B
& 25.0 & 91.6 & 22.0 & 95.0 & 0.6 & 3.8 & 74.4 & 4.6 \\

\cmidrule(lr){3-11}

& & GPT-4o
& 94.6 & 94.4 & 96.8 & 97.8 & 3.2 & 3.8 & 2.2 & 1.8 \\

& & DeepSeek-v3.1
& 94.8 & 95.8 & 98.6 & 99.6 & 4.0 & 4.0 & 1.2 & 0.2 \\

& & Claude Sonnet 4.5
& 60.2 & 57.2 & 58.0 & 54.6 & 1.0 & 0.8 & 38.8 & 42.0 \\

\bottomrule
\end{tabular}

\label{tab:Qwen3-32B_judge}
\end{table*}


\begin{table*}[t]
\centering
\caption{Evaluation results(\%) of LLM-as-a-Judge behavior with and without reasoning chains across factual and mathematical datasets, with all answers generated by DeepSeek-v3.1.}
\scriptsize

\setlength{\tabcolsep}{6pt}
\begin{tabular}{l c l cc cc cc cc}
\toprule
\multirow{2}{*}{Dataset}
& \multirow{2}{*}{Acc}
& \multirow{2}{*}{Judge Models}
& \multicolumn{2}{c}{Alignment}
& \multicolumn{2}{c}{Pass Rate}
& \multicolumn{2}{c}{Overconfidence}
& \multicolumn{2}{c}{Conservativeness} \\
\cmidrule(lr){4-5} \cmidrule(lr){6-7} \cmidrule(lr){8-9} \cmidrule(lr){10-11}
 &  & 
& w/o Think & w/ Think
& w/o Think & w/ Think
& w/o Think & w/ Think
& w/o Think & w/ Think \\
\midrule

\multirow{6}{*}{NQ}
& \multirow{6}{*}{46.4}
& Qwen3-8B
& 75.8 & 50.6 & 46.6 & 94.2 & 12.2 & 48.6 & \textbf{12.0} & 0.8 \\

& & Qwen3-14B
& 75.8 & 49.0 & 54.2 & 95.4 & 16.0 & 50.0 & 8.2 & 1.0 \\

& & Llama3-8B
& 77.8 & 47.0 & \textbf{61.0} & \textbf{98.6} & \textbf{18.4} & \textbf{52.6} & 3.8 & 0.4 \\

\cmidrule(lr){3-11}

& & GPT-4o
& 88.0 & 51.6 & 52.8 & 91.6 & 9.2 & 46.8 & 2.8 & 1.6 \\

& & DeepSeek-V3.1
& 80.4 & 49.0 & 46.0 & 95.8 & 9.6 & 50.2 & 10.0 & 0.8 \\

& & Claude Sonnet 4.5
& \textbf{90.0} & \textbf{55.8} & 48.8 & 69.8 & 6.2 & 33.8 & 3.8 & \textbf{10.4} \\

\midrule
\multirow{6}{*}{GSM8K}
& \multirow{6}{*}{95.6}
& Qwen3-8B
& 56.0 & 92.6 & 53.2 & 95.8 & 0.8 & 3.8 & 43.2 & 3.6 \\

& & Qwen3-14B
& 63.4 & 94.4 & 61.8 & 97.2 & 1.4 & 3.6 & 35.2 & 2.0 \\

& & Llama3-8B
& 51.8 & 94.2 & 51.8 & 97.8 & \textbf{2.2} & \textbf{4.0} & \textbf{46.0} & 1.8 \\

\cmidrule(lr){3-11}

& &GPT-4o
& \textbf{66.4} & \textbf{95.0} & \textbf{64.0} & 97.0 & 1.0 & 3.2 & 32.6 & 1.8 \\

& & DeepSeek-v3.1
& 64.0 & \textbf{95.0} & \textbf{64.0} & \textbf{98.2} & \textbf{2.2} & 3.8 & 33.8 & 1.2 \\

& & Claude Sonnet 4.5
& 58.6 & 87.6 & 55.4 & 85.6 & 0.6 & 1.2 & 40.8 & \textbf{11.2} \\

\bottomrule
\end{tabular}

\label{tab:DeepSeek-v3.1_judge}
\end{table*}

\begin{table*}[t]
\centering
\caption{Evaluation results (\%) of LLM-as-a-Judge behavior on MATH500, with all answers generated by Qwen3-8B.}

\scriptsize
\setlength{\tabcolsep}{6pt}
\renewcommand{\arraystretch}{1.15}
\begin{tabular}{lllcccccccc}
\toprule
Dataset & Acc & Judge Models 
& \multicolumn{2}{c}{Alignment} 
& \multicolumn{2}{c}{Pass Rate} 
& \multicolumn{2}{c}{Overconfidence} 
& \multicolumn{2}{c}{Conservativeness} \\
\cmidrule(lr){4-5} \cmidrule(lr){6-7} \cmidrule(lr){8-9} \cmidrule(lr){10-11}
& & 
& w/o Think & w/ Think 
& w/o Think & w/ Think 
& w/o Think & w/ Think 
& w/o Think & w/ Think \\
\midrule

\multirow{6}{*}{MATH500} 
& \multirow{6}{*}{84.8}
& Qwen3-8B          & 85.0 & 86.6 & 86.6 & 97.8 & 8.4  & 13.2 & 6.6  & 0.2 \\
& & Qwen3-14B        & 88.6 & 86.8 & 84.2 & 98.0 & 5.4  & 13.2 & 6.0  & 0.0 \\
& & Llama3-8B        & 88.6 & 88.8 & 92.2 & 83.6 & 9.4  & 5.0  & 2.0  & 6.2 \\
\cmidrule(lr){3-11}
& & DeepSeek-v3.1    & 86.0 & 85.8 & 98.4 & 98.2 & 13.8 & 13.8 & 0.2  & 0.4 \\
& & GPT-4o            & 86.8 & 88.8 & 74.0 & 76.4 & 1.2  & 1.4  & 12.0 & 9.8 \\
& & Claude Sonnet 4.5 & 77.8 & 89.4 & 94.6 & 76.6 & 10.2 & 7.0  & 0.4  & 15.2 \\

\bottomrule
\end{tabular}

\label{tab:math500_balanced_results}
\end{table*}

\begin{table*}[t]
\centering
\caption{Evaluation results (\%) of LLM-as-a-Judge behavior on a fully balanced subset (1:1 ratio of correct and incorrect samples), constructed by pairing 70 correct answers with 70 randomly selected incorrect answers.}

\scriptsize
\setlength{\tabcolsep}{6pt}
\renewcommand{\arraystretch}{1.15}
\begin{tabular}{lllcccccccc}
\toprule
Dataset & Acc & Judge Models 
& \multicolumn{2}{c}{Alignment} 
& \multicolumn{2}{c}{Pass Rate} 
& \multicolumn{2}{c}{Overconfidence} 
& \multicolumn{2}{c}{Conservativeness} \\
\cmidrule(lr){4-5} \cmidrule(lr){6-7} \cmidrule(lr){8-9} \cmidrule(lr){10-11}
& & 
& w/o Think & w/ Think 
& w/o Think & w/ Think 
& w/o Think & w/ Think 
& w/o Think & w/ Think \\
\midrule

\multirow{6}{*}{MATH500} 
& \multirow{6}{*}{50.0}

& Qwen3-8B & 67.9 & 57.9 & 70.7 & 92.1 & 26.4 & 42.1 & 5.7 & 0.0 \\
& & Qwen3-14B & 80.0 & 56.4 & 64.3 & 93.6 & 17.1 & 43.6 & 2.9 & 0.0 \\
& & Llama3-8B & 52.9 & 63.6 & 77.1 & 83.6 & 37.1 & 35.0 & 10.0 & 1.4 \\
\cmidrule(lr){3-11}
& & DeepSeek-v3.1 & 58.6 & 56.4 & 91.4 & 93.6 & 41.4 & 43.6 & 0.0 & 0.0 \\
& & GPT-4o & 95.7 & 90.0 & 47.1 & 48.6 & 0.7 & 4.3 & 3.6 & 5.7 \\
& & Claude Sonnet 4.5 & 63.6 & 62.1 & 83.6 & 66.4 & 35.0 & 27.1 & 1.4 & 10.7 \\

\bottomrule
\end{tabular}

\label{tab:math500_results}
\end{table*}

\begin{table*}[t]
\centering
\small
\setlength{\tabcolsep}{6pt}
\renewcommand{\arraystretch}{1.15}
\caption{Judge pass rates (\%) on the NQ dataset under progressively injected erroneous sentences in reasoning chains.}
\label{tab:fine_grained_errors}

\begin{tabular}{lccccccc}
\toprule
Judge Models
& \multicolumn{2}{c}{Vanilla}
& Basic-All
& Wrong-1
& Wrong-2
& Wrong-3
& Wrong-All \\

\cmidrule(lr){2-3}

& w/o Think & w/ Think
& w/ Think
& w/ Think
& w/ Think
& w/ Think
& w/ Think \\

\midrule

Qwen3-8B & 63.0 & 91.0 & 61.0 & 58.8 & 51.4 & 52.2 & 47.6 \\
Qwen3-14B & 62.8 & 87.4 & 58.0 & 55.0 & 51.4 & 52.2 & 53.2 \\
Llama3-8B & 74.8 & 97.2 & 92.8 & 95.0 & 88.6 & 87.6 & 95.2 \\
GPT-4o & 50.6 & 48.6 & 18.4 & 13.4 & 8.6 & 7.8 & 23.4 \\
DeepSeek-v3.1 & 45.2 & 24.4 & 1.0 & 1.0 & 0.0 & 0.0 & 0.0 \\

\bottomrule
\end{tabular}

\end{table*}

\end{document}